\pdfoutput=1
\documentclass[11pt,journal]{IEEEtran}

\usepackage{amsmath,amsfonts}
\usepackage{algorithmic}
\usepackage{algorithm}
\usepackage{array}
\usepackage[caption=false,font=normalsize,labelfont=sf,textfont=sf]{subfig}
\usepackage{textcomp}
\usepackage{stfloats}
\usepackage{url}
\usepackage{verbatim}
\usepackage{graphicx}
\usepackage{xcolor}
\usepackage{cite}
\usepackage{hyperref}
\usepackage{booktabs}
\usepackage{cleveref}
\usepackage{comment}
\usepackage{booktabs}
\usepackage{tabularx}
\usepackage{float}
\usepackage{array}
\usepackage{makecell}
\usepackage{float}
\usepackage{array}
\usepackage{graphicx}
\usepackage{wrapfig}
\newcolumntype{C}[1]{>{\centering\arraybackslash}p{#1}}
\newcolumntype{J}[1]{>{\raggedright\arraybackslash}p{#1}}
\begin{document}

\title{Implicit Neural Representations: A Signal Processing Perspective}

\author{Dhananjaya Jayasundara and Vishal M.~Patel%
\thanks{Dhananjaya Jayasundara and Vishal M.~Patel are with the Department of Electrical and Computer Engineering, Johns Hopkins University, Baltimore, MD 21218 USA (Corresponding author: vjayasu1@jhu.edu).}}

\markboth{}%
{Jayasundara and Patel: Implicit Neural Representations: A Signal Processing Perspective}

\maketitle


\section{Introduction}
\label{sec:introduction}

Signals are traditionally represented through discrete samples, transform coefficients, or other grid-based descriptions. Images are stored on pixel lattices, audio as sampled waveforms, videos as sequences of frames, and volumetric data as voxels. This discrete view has been extraordinarily successful, forming the basis of modern sensing, compression, storage, and reconstruction pipelines \cite{smith1997scientist}. In such pipelines, a continuous physical phenomenon is first sampled, then processed and stored in discrete form, often after transformation into a domain that exposes structure such as sparsity or energy compaction. Despite this success, many tasks of interest in signal and image processing are inherently continuous in nature \cite{sitzmann2020implicit}. Interpolation, super-resolution, geometric warping, and inverse reconstruction all require reasoning about values between samples or beyond the observed domain. In conventional approaches, this continuous behavior is introduced only after discretization, typically through interpolation kernels, regularization models, or iterative reconstruction procedures. As a result, the representation itself remains fundamentally tied to a fixed sampling lattice, and the notion of continuity is external to the model rather than intrinsic to it.

Implicit neural representations (INRs) offer a different perspective. Instead of representing a signal as a collection of discrete samples, they model it as a continuous function parameterized by a neural network,
\[
f_{\boldsymbol{\theta}} : \Omega \subset \mathbb{R}^{d} \rightarrow \mathbb{R}^{c},
\]
where \(\Omega\) denotes the signal domain (e.g., spatial coordinates \(x \in \mathbb{R}^2\) for images, spatio-temporal coordinates \((x,t) \in \mathbb{R}^3\) for videos, or spatial-directional coordinates for radiance fields), and \(c\) is the number of signal channels (e.g., \(c=1\) for grayscale, \(c=3\) for RGB). The parameter vector \(\boldsymbol{\theta} \in \mathbb{R}^{p}\) encodes the signal implicitly through the network weights. In this formulation, querying the signal at any location reduces to evaluating the network at the corresponding coordinate. This seemingly simple idea shifts the role of representation: the signal is no longer an array of values, but a function that can be evaluated continuously over its domain.

\begin{figure*}[t]
    \centering
    \includegraphics[width=\textwidth]{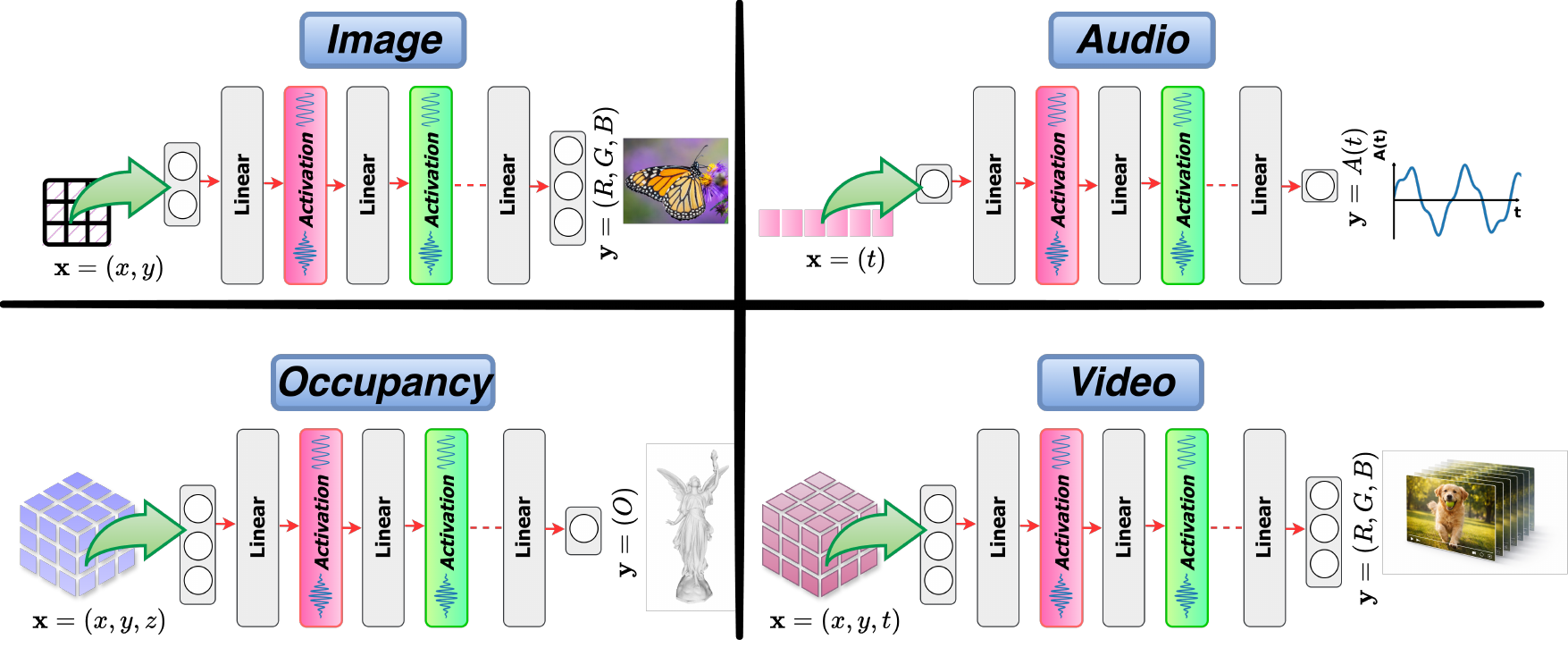}
    \caption{A unified view of INRs across signal modalities. The same coordinate-based neural model can represent images, audio, occupancy fields, and videos by changing only the input domain and output quantity. INRs thus provide a modality-agnostic framework for modeling diverse signals.}
    \label{fig:main_img}
\end{figure*}

A notable consequence of this formulation is its generality. By changing only the domain \(\Omega\) and the output quantity, the same coordinate-based model can represent images, audio, video, geometry, and related modalities. As illustrated in \Cref{fig:main_img}, images correspond to mappings from spatial coordinates to RGB values, audio to mappings from time to amplitude, geometric representations such as occupancy fields or signed distance functions to mappings from spatial coordinates to structural properties, and video corresponds to mapping spatio-temporal coordinates to RGB values (see \Cref{sec:video} for efficient video architectures). This unified view highlights a key conceptual shift, rather than designing separate representations for each modality, INRs provide a common functional framework that adapts naturally to the signal domain.

\begin{figure*}[t]
    \centering
    \includegraphics[width=\textwidth]{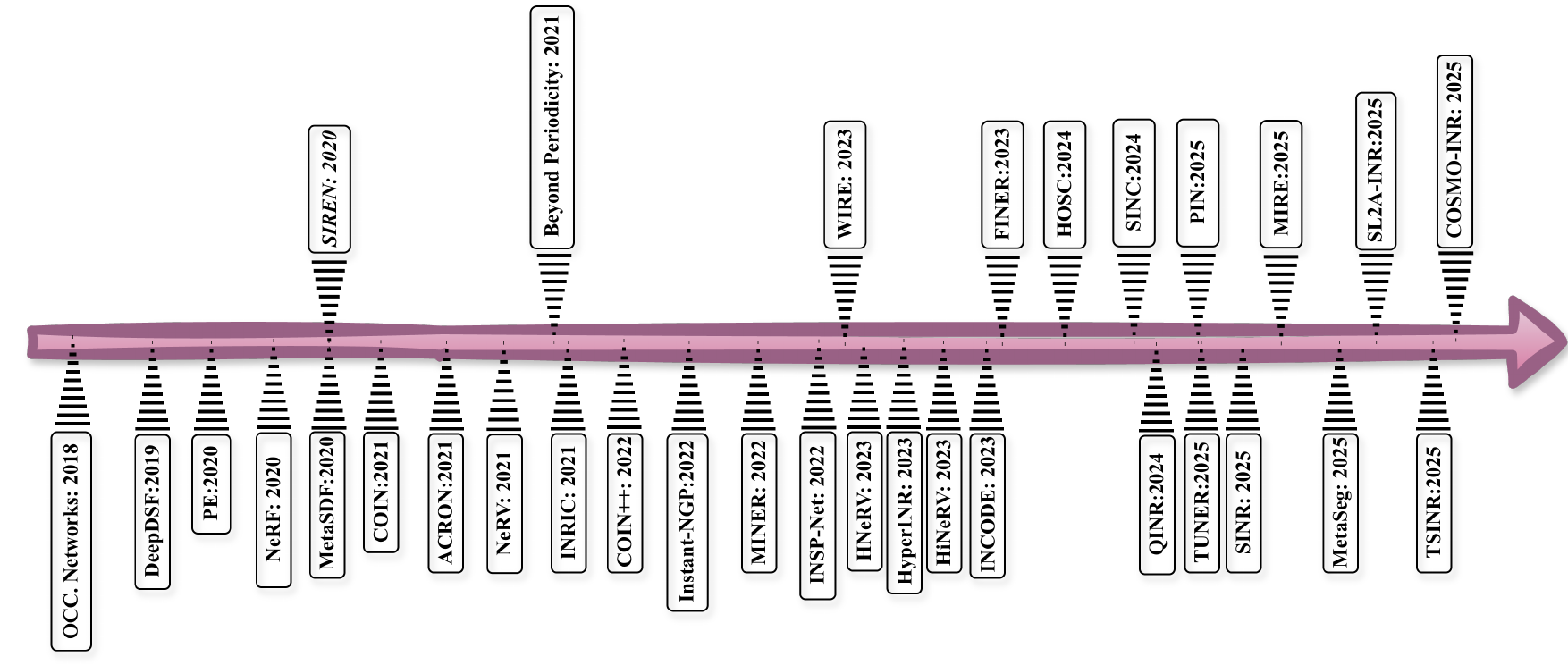}
    \caption{Evolution of INRs. A chronological view of key developments, illustrating the transition from early coordinate-based models to more advanced architectures emphasizing frequency modeling, multiscale structure, and signal-adaptive design.}
    \label{fig:progression}
\end{figure*}

This shift from \emph{samples} to \emph{functions} is not merely a change in implementation; it changes the object being modeled. A pixel grid is only a discrete realization of an underlying visual signal, tied to a particular resolution and sampling pattern. An INR instead seeks to represent the underlying signal itself, independent of how it is sampled. This distinction becomes particularly important in settings where the signal is observed indirectly, sparsely, or at varying resolutions. For instance, view-dependent appearance, continuous geometry, and spatially varying physical quantities are often cumbersome to model with fixed discrete structures, requiring large memory or ad hoc interpolation schemes. A representation that is continuous from the outset is often more faithful to the structure of such problems. At the same time, this functional viewpoint raises fundamental questions that are deeply familiar to signal processing. For instance,  How does the representation handle different spectral components of the signal, particularly fine-scale detail? What role do the choice of coordinate encoding, activation functions, and network architecture play in shaping the class of functions that can be represented? These questions are not new; they echo classical concerns about sampling, reconstruction, approximation spaces, and basis design. However, in the INR setting, they reappear in a learned and data-adaptive form. One of the most immediate aspects where this connection becomes evident is spectral behavior. Neural networks are known to exhibit a preference for low-frequency structure, often capturing coarse components of a signal before refining high-frequency details \cite{rahaman2019spectral, tancik2020fourier, saragadam2023wire, ramasinghe2022beyond}. This phenomenon, commonly referred to as spectral bias, has important implications for INR design. Many developments in the field can be interpreted as attempts to address this limitation, by modifying coordinate encodings, introducing periodic or localized activation functions, or incorporating multiscale structure. When we look at these approaches from a signal processing perspective, it becomes clear that they shape the underlying approximation space, much like choosing an appropriate basis or transform in classical representations. \Cref{fig:progression} summarizes this evolution of INR models, illustrating the field’s ongoing attempt to reconcile expressivity, locality, and efficiency in continuous signal representations.

\vspace{-3pt}

Another important consequence of representing signals as continuous neural functions is that many signal operations can be defined directly on the representation itself. As \(f_{\theta}\) is differentiable by construction, quantities such as gradients, Laplacians, and integrals can be computed analytically or through automatic differentiation, rather than approximated through discrete operators \cite{sitzmann2020implicit}. This is particularly appealing in applications involving physical fields, geometry, and scientific data, where differential relationships and measurement consistency play a central role. In this sense, INRs blur the boundary between representation and computation, enabling a more unified treatment of signals and the operators acting on them. Finally, the shift to continuous neural representations also has implications for how signal complexity is modeled. Natural signals are inherently non-uniform, containing both smooth regions and fine-scale structure. Classical signal processing addresses this through multiscale and localized representations such as wavelets and pyramids. Recent INR approaches similarly recognize that a single global function may not be sufficient to capture all scales of variation effectively, motivating the development of multiresolution encodings and locally adaptive mechanisms \cite{saragadam2022miner, liu2024finer}. These developments further reinforce the view that INRs are best understood not as monolithic models, but as flexible function spaces whose properties depend on how they are constructed and trained.

This article adopts a signal-processing perspective on INRs by design. Rather than cataloging architectures, we interpret INRs through the fundamental lenses of spectral behavior, sampling and reconstruction, and multiscale representation. This perspective situates INRs within a broader signal-processing lineage, clarifies the key challenges that remain, and provides a natural framework for discussing their broad range of applications across diverse domains. By framing INRs as continuous signal models shaped by architectural and representational choices, we aim to provide both intuition and rigor, and to highlight why they have emerged as a compelling paradigm for modern signal and image processing.

\section{INR Fundamentals and Signal-Processing Formulation}
\label{sec:fundamentals}

Having developed the basic perspective on INRs, we now make the underlying framework more explicit and precise. The goal of this section is to fix notation, clarify the underlying learning problem, and identify the main components that shape representational behavior. Rather than surveying architectural variants, we focus on the minimal formulation needed for the rest of the article,  what an INR represents, how it is learned from data, and which design choices influence its approximation, regularity, and operator-related properties.
In general, an INR is modeled as a parametric continuous function
\begin{equation}
f_{\theta} : \Omega \subset \mathbb{R}^{d} \rightarrow \mathbb{R}^{c},
\label{eq:inr_def}
\end{equation}
where $\mathbf{x}\in\Omega$ denotes a coordinate in the underlying domain, $f_{\theta}(\mathbf{x})$ is the corresponding signal value, and $\theta$ denotes the learnable parameters of the neural model. The domain $\Omega$ may represent space, time, direction, space-time, or other acquisition variables, depending on the signal of interest. The output dimension $c$ is likewise modality dependent. In this formulation, the neural network is not merely a predictor applied to a signal representation; it \emph{is} the representation.
At first glance, \Cref{eq:inr_def} resembles a standard regression model. However, the interpretation is fundamentally different. In conventional supervised learning, the network is often viewed as a predictor that maps input data to labels or outputs of downstream interest. In the INR setting, the network itself is the representation of the underlying signal. Its parameters are not merely auxiliary variables for optimization; they are the signal code. We believe this distinction is crucial as many properties that appear unusual from a standard deep-learning viewpoint become natural once the neural network is regarded as a continuous signal model.

\subsection{From discrete samples to continuous neural functions}
\label{subsec:discrete_to_continuous}

In the INR setting, the observed discrete samples are treated not as the representation itself, but as measurements of an underlying continuous signal. This shifts the role of the model: rather than storing the signal directly on a fixed lattice, the goal is to infer a continuous function that can be queried at arbitrary coordinates. Classical reconstruction methods achieve this through predefined models such as splines, interpolation kernels, or transform-domain synthesis \cite{smith1997scientist}. INRs instead learn this reconstruction map directly through a parametric neural function. Let \(s:\Omega \rightarrow \mathbb{R}^{c}\) denote the unknown signal defined on a continuous domain \(\Omega \subset \mathbb{R}^{d}\), and let \(\{(\mathbf{x}_i,\mathbf{y}_i)\}_{i=1}^{N}\) denote the observed samples, where \(\mathbf{x}_i \in \Omega\) is the coordinate of the \(i\)-th sample and \(\mathbf{y}_i \in \mathbb{R}^{c}\) is the corresponding measured value. Then the sampled-data model can be written as
\(
\mathbf{y}_i \approx s(\mathbf{x}_i), \qquad \hat{s}(\mathbf{x}) = f_{\theta}(\mathbf{x})
\) where \(f_{\theta}\) is the INR and \(\hat{s}\) denotes the learned approximation to the underlying signal \(s\). In this view, the discrete measurements are not themselves the representation; rather, they are observations from which a continuous representation is inferred. This viewpoint is closely connected to classical approximation theory in signal processing. A standard approach is to approximate \(s\) in a prescribed basis or dictionary:
\(s(\mathbf{x}) \approx \sum_{k=1}^{K} a_k \,\phi_k(\mathbf{x})\), 
where \(\{\phi_k\}_{k=1}^{K}\) are fixed atoms, such as Fourier basis functions, wavelets, splines, or radial basis functions, and \(\{a_k\}_{k=1}^{K}\) are the associated coefficients. The approximation problem is then reduced to estimating the coefficient vector \(\mathbf{a}=(a_1,\dots,a_K)\). From this perspective, one may interpret an INR as replacing a fixed linear approximation family by a learned nonlinear function class \(\mathcal{F}_{\Theta}=\{f_{\theta}:\theta\in\Theta\}\),
whose elements are determined jointly by the architecture, coordinate encoding, activation functions, and network parameters. The representation remains parametric, but the approximation space is no longer specified explicitly through a predetermined basis. Instead, the admissible family of approximants is induced implicitly by the design of the neural model itself. A precise distinction lies in how these methods expose their representational components. Classical approximation frameworks usually maintain an explicit separation: in fixed-basis models, atoms are predetermined and only coefficients are estimated; in adaptive constructions such as dictionary learning \cite{jayasundara2025sinr}, both atoms and coefficients are learned but remain conceptually distinct. In both cases, the signal is reconstructed via an explicit linear combination. INRs diverge from this paradigm by rendering this separation \emph{implicit}. The effective representational atoms are neither predefined nor cleanly disentangled from the parameters that govern them. Instead, they emerge from the non-linear interaction of coordinate encodings, architectural topology, and learned weights.

This is what makes INRs both powerful and challenging from a signal-processing viewpoint. Their appeal lies in flexibility: they define a learned nonlinear approximation space that can adapt to the structure of the target signal rather than committing to a fixed linear expansion. But that same flexibility also obscures many of the properties that are easier to interpret in classical settings, such as bandwidth, locality, approximation order, sparsity, and stability \cite{smith1997scientist}. Those properties are no longer read directly from an explicit transform domain; they must instead be understood through the joint effect of model design and  \cite{jayasundara2025sinr}. In this sense, INRs do not replace approximation theory so much as expand it into a setting where the approximation space itself is part of what must be learned and analyzed. Additionally, once the function $f_{\theta}$ is learned, the signal can be queried on any grid, not only on the grid used during training \cite{sitzmann2020implicit}. In principle, this allows arbitrary resampling, super-resolution, and nonuniform querying \cite{saragadam2023wire}. In practice, the quality of such querying depends on the learned function class, the available measurements, and the spectral content of the underlying signal \cite{li2023regularize, saragadam2023wire, xu2022signal}. Despite the many factors that affect final quality, the main conceptual appeal remains that resolution is separated from the representation itself.

\subsection{Canonical learning objectives for INRs}
\label{subsec:learning_objectives}

The most basic INR learning problem is direct function fitting from samples, as introduced in Section~\ref{sec:introduction}. Given coordinates \(\{\mathbf{x}_i\}_{i=1}^{N}\) and corresponding values \(\{\mathbf{y}_i\}_{i=1}^{N}\), one estimates \(\theta\) by solving
\begin{equation}
\min_{\theta} \; \frac{1}{N}\sum_{i=1}^{N} \ell\!\left(f_{\theta}(\mathbf{x}_i), \mathbf{y}_i\right),
\label{eq:basic_empirical_risk_rewrite}
\end{equation}
where \(\ell\) is typically a pointwise distortion such as squared error, absolute error, or another task-dependent discrepancy \cite{sitzmann2020implicit, saragadam2023wire}. In many early INR demonstrations, this direct overfitting formulation was sufficient: one network was trained for one signal in order to study representational fidelity, continuity, or derivative accuracy \cite{sitzmann2020implicit}. Even this simple objective already admits a signal-processing interpretation. First, it may be viewed as nonlinear approximation from sampled data. Second, when the observations are noisy,  \Cref{eq:basic_empirical_risk_rewrite} behaves like a denoising or smoothing estimator whose behavior is governed not only by the loss, but also by the implicit bias of the model class \cite{rahaman2019spectral, saragadam2023wire}. Third, the sampling geometry itself matters: whether the observations are uniform, nonuniform, sparse, or importance-weighted directly determines which parts of the signal are well constrained \cite{li2023regularize, saragadam2023wire}. Even though \Cref{eq:basic_empirical_risk_rewrite} resembles ordinary regression, its meaning is closer to estimation under a continuous signal model than to generic prediction. In many important applications; however, one does not observe the signal values directly. Instead, one observes measurements generated through an acquisition or physical measurement process. This is naturally expressed through a forward operator \(\mathcal{A}\), leading to the more general objective
\begin{equation}
\min_{\theta} \; \mathcal{D}\!\left(\mathcal{A}(f_{\theta}), \mathbf{y}\right) + \lambda \mathcal{R}(\theta,f_{\theta}),
\label{eq:inverse_problem_inr_rewrite}
\end{equation}
where \(\mathbf{y}\) denotes the measurements, \(\mathcal{D}\) is a data-fidelity term, and \(\mathcal{R}\) is a regularizer or prior. This formulation is particularly important because it makes explicit that an INR is not merely a coordinate-to-value regressor; it is a latent signal model embedded inside an observation model. The role of \(\mathcal{A}\) is therefore central. It attempts to capture how the underlying continuous signal is seen by the measurement process, whether through sampling, masking, convolution, projection, integration along rays, or differential constraints. In this sense, \(\mathcal{A}\) is the bridge between the continuous representation and the observed data. This perspective has several important implications. First, the recovery problem is generally no longer a pure interpolation problem, but an inverse problem: the network must infer a signal that is consistent with observations only after they have passed through \(\mathcal{A}\). Second, identifiability is now determined not only by the expressivity of the INR, but also by the information preserved or discarded by the forward model. If \(\mathcal{A}\) is lossy, ill-conditioned, or incomplete, then many distinct continuous signals may explain the same measurements. Third, the forward operator determines which priors are meaningful. For example, a sampling or masking operator naturally gives rise to inpainting or sparse interpolation; a blur or convolution operator leads to deblurring; projection operators lead to tomographic reconstruction; rendering operators lead to neural field estimation from images; and differential operators lead to PDE-constrained fitting. Therefore,  \Cref{eq:inverse_problem_inr_rewrite} is not just a generalization of  \Cref{eq:basic_empirical_risk_rewrite}; it is the formulation that makes the signal-processing content of INR learning explicit. The most widely used data-fidelity term for direct fitting is the mean squared error (MSE) \cite{sitzmann2020implicit,ramasinghe2022beyond,saragadam2022miner,saragadam2023wire,liu2024finer, li2023regularize, jayasundara2025pin}
\begin{equation}
\ell_{\text{MSE}} = \frac{1}{N}\sum_{i=1}^{N} \|f_{\theta}(\mathbf{x}_i)-\mathbf{y}_i\|_2^2.
\label{eq:mse_loss_rewrite}
\end{equation}
This choice is natural when the measurement noise is modeled as approximately Gaussian, as minimizing the MSE
error then corresponds to maximum-likelihood estimation, and it is also consistent with settings where one seeks high peak signal-to-noise ratio (PSNR). In other settings, absolute error or robust losses may be preferable, depending on the application \cite{liu2025stereoinr}. For perceptual signals such as images, audio, or video, one may also introduce task-aware or perceptual discrepancies in order to align reconstruction quality with downstream objectives or human perception \cite{chen2021nerv}. In the inverse-problem setting, however, the fidelity term often acts not on the signal itself, but on \(\mathcal{A}(f_{\theta})\), which means that the notion of reconstruction accuracy is tied directly to the sensing process.

A distinctive feature of INRs is that the learned signal is differentiable with respect to its input coordinates, provided the activation function is sufficiently smooth \cite{sitzmann2020implicit,xu2022signal}. This makes it possible to impose supervision not only on signal values, but also on derivatives and more general differential operators. For example, if derivative information is available or prescribed, one may impose a gradient-domain loss of the form
\(\mathcal{L}_{\nabla}
=
\frac{1}{N}\sum_{i=1}^{N}
\left\|
\nabla_{\mathbf{x}} f_{\theta}(\mathbf{x}_i)
-
\mathbf{g}_i
\right\|_2^2,\) where \(\mathbf{g}_i\) denotes the target gradient of the signal at coordinate \(\mathbf{x}_i\). Likewise, if higher-order differential information is relevant, one may impose an operator-domain loss such as \(\mathcal{L}_{\Delta}
=
\frac{1}{N}\sum_{i=1}^{N}
\left\|
\Delta f_{\theta}(\mathbf{x}_i)
-
h_i
\right\|_2^2,\) where \(h_i\) denotes the target value of the Laplacian at \(\mathbf{x}_i\). This is one of the most important ways in which INRs differ from conventional discrete representations. Because the signal is modeled as a continuous function rather than merely as a sampled array, derivative-domain and operator-domain constraints can be imposed analytically through automatic differentiation, rather than being approximated solely through finite-difference stencils. In practical terms, this means that the forward operator \(\mathcal{A}\) in \Cref{eq:inverse_problem_inr_rewrite} may itself involve continuous differential structure, not just discrete measurements. At the same time, such derivative-based supervision is not free: computing derivatives of the network output with respect to the input coordinates, especially higher-order derivatives, can be computationally and memory intensive.

In \Cref{eq:inverse_problem_inr_rewrite}, the regularizer \(\mathcal{R}\) may take several forms. The simplest are parameter penalties such as weight decay, \(\mathcal{R}_{\theta} = \|\theta\|_2^2\), but more structured regularizers may enforce smoothness, sparsity, derivative consistency, total variation, spectral concentration, or physical constraints \cite{xu2022signal}. Crucially, regularization in INR models is not limited to explicit penalty terms. It also enters implicitly through the architecture, coordinate encoding, activation family, and optimization dynamics \cite{li2023regularize}. This is especially important when \(\mathcal{A}\) is ill-posed or incomplete, as in such cases the effective prior induced by the representation may matter as much as the explicit loss itself. Finally, the optimization in \Cref{eq:basic_empirical_risk_rewrite} and \Cref{eq:inverse_problem_inr_rewrite} is often performed for a single signal. That is, one network is trained per image, audio clip, volume, or scene. This provides a conceptually transparent view of signal representation, directly exposing the model’s approximation properties and its implicit regularization behavior. An alternative is \emph{amortized} or \emph{conditional} INR learning, in which a shared model or hypernetwork maps an observed signal, latent code, or side information to the parameters of an INR. This shifts the problem from estimating one continuous function at a time to learning a prior over a family of such functions (see \Cref{sec:efficient}).

\subsection{Spectral design in coordinate-based models.}
\begin{figure}[t]
    \centering
    \includegraphics[width=\linewidth]{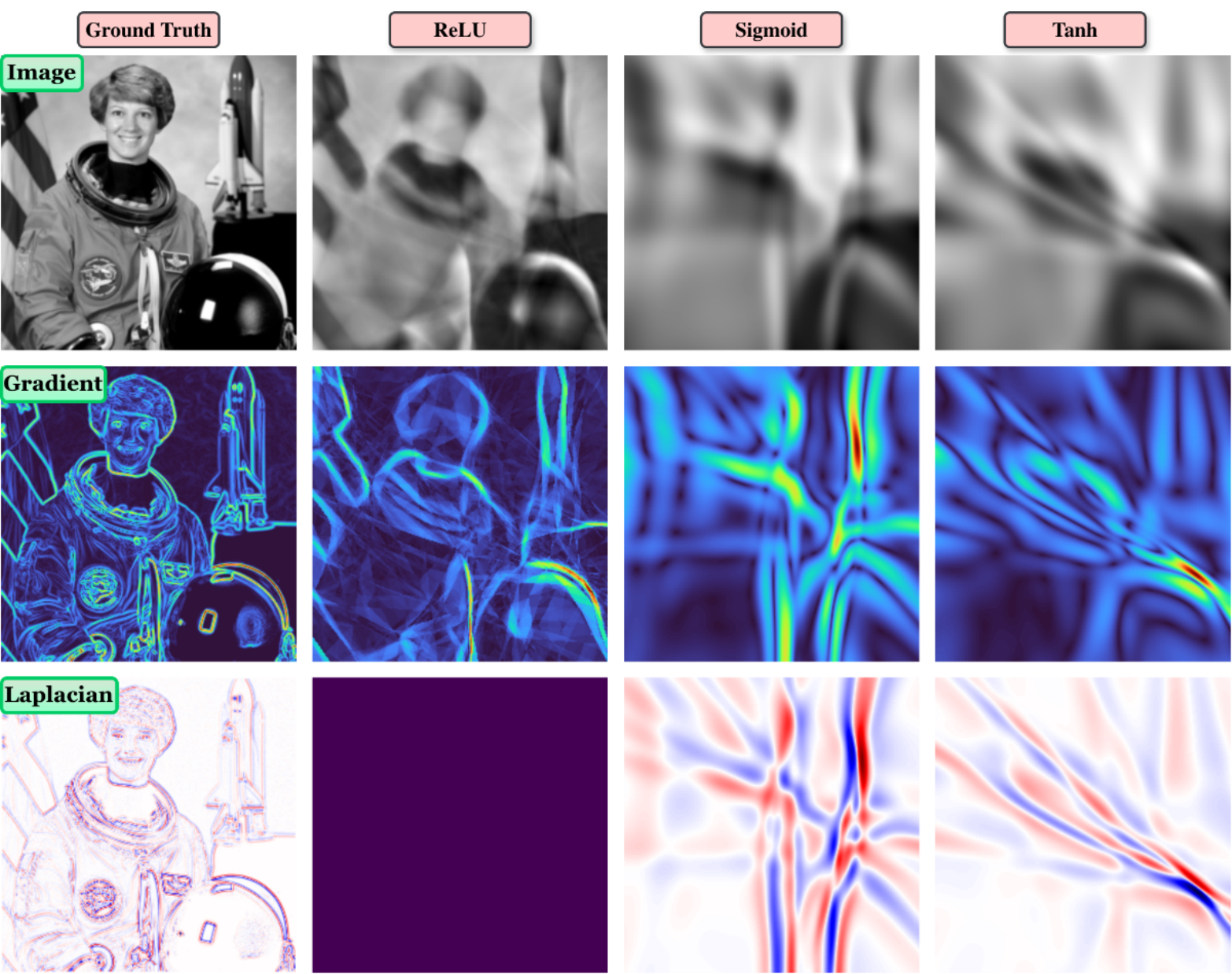}
    \caption{Failure of standard activations to preserve signal and operator fidelity in coordinate-based models. Beyond the oversmoothed reconstruction, larger discrepancies appear in the gradient and Laplacian fields, where edges and second-order variations are distorted, revealing spectral bias beyond reconstruction.}
    \label{fig:standard_activations_failure}
\end{figure}

A useful way to organize the development of INR methodology is to begin with the central limitation of vanilla coordinate-based multilayer-perceptrons (MLPs): \emph{spectral bias} \cite{rahaman2019spectral}. When conventional activations such as ReLU, Sigmoid, and Tanh are employed in coordinate-based networks, the resulting models tend to favor smooth, low-frequency structure and recover fine detail only slowly \cite{sitzmann2020implicit,tancik2020fourier, saragadam2023wire}. In the INR setting, this is not merely an optimization inconvenience; it means that the induced approximation space is spectrally unbalanced. Coarse components are represented efficiently, whereas rapidly varying or highly oscillatory structure is poorly conditioned. Fig.~\ref{fig:standard_activations_failure} shows visual evidence for this spectral biased scenario. With conventional activations, the reconstruction becomes visibly oversmoothed, but as can be seen from \Cref{fig:standard_activations_failure} the discrepancy is even more pronounced in the gradient and Laplacian fields, where edge structure and second-order variation are strongly distorted. 
This is particularly important for INRs, as many scientific and engineering downstream tasks may depend not only on function values, but also on accurate first- and second-order operators. Spectral bias is therefore best understood not simply as a reconstruction artifact, but as a deeper limitation in the differential fidelity of the learned representation. The developments that followed can be understood as a progression in how this approximation space is reshaped: first through changes to the input representation, then through modifications to the activation and initialization, next through changes in the training parameterization, and finally through a shift toward localized, adaptive, and sampling-theoretically grounded atoms.

\subsubsection{\textcolor{blue}{Lifting Coordinates into the Frequency Domain}} A natural first response to spectral bias is to modify not the network itself, but the way coordinates are presented to it. This is the role of \emph{positional encoding or frequency encoding} (PE). Instead of feeding the raw coordinate \(\mathbf{x}\in\mathbb{R}^{d}\) directly into the neural network, one first maps it into a higher-dimensional feature space whose components already contain some oscillatory structure \cite{tancik2020fourier}.  This modified coordinates to the network can be written as
\(\gamma_{\mathbf{B}}(\mathbf{x})
=
\big[
\cos(2\pi \mathbf{B}\mathbf{x}),
\;
\sin(2\pi \mathbf{B}\mathbf{x})
\big]\), 
where \(\mathbf{B}\in\mathbb{R}^{m\times d}\) is a matrix of frequencies. As \(\mathbf{B}\mathbf{x}\in\mathbb{R}^{m}\), the encoded coordinate \(\gamma_{\mathbf{B}}(\mathbf{x})\) lies in \(\mathbb{R}^{2m}\): each of the \(m\) projected frequencies contributes one cosine and one sine component. Accordingly, the formulation in \Cref{eq:inr_def} is modified slightly to accept the lifted high-dimensional coordinates. The INR then takes the form \( f_{\theta}(\mathbf{x}) = g_{\theta}\!\big(\gamma_{\mathbf{B}}(\mathbf{x})\big) \), where \(g_{\theta}\) acts on harmonic coordinate features instead of only the original spatial or temporal coordinates. This modification changes the role of the network. Rather than relying entirely on its hidden layers to build oscillatory structure from low-dimensional inputs, the model is provided with input features that already span a richer set of frequencies. The network therefore learns how to combine these harmonic components to reconstruct fine-scale variation more effectively. Positional encoding alters the effective approximation space prior to learning: high-frequency components are explicitly introduced at the input, and the network primarily learns how to combine them. In this sense, the encoding can be interpreted as a predetermined harmonic dictionary over the coordinate domain. This stage is analogous to a modulation or feature-lifting operation that expands the spectral support accessible to the downstream network.
The choice of the frequency matrix \(\mathbf{B}\) in the mapping \(\gamma_{\mathbf{B}}(\mathbf{x})\) plays a central role in determining the representation capacity of the resulting INR.  

This construction, \(\mathbf{B}\) directly controls which harmonic components are exposed to the network at the input stage. A common choice is to sample \(\mathbf{B}\) from a Gaussian distribution, i.e., \(\mathbf{B}_{ij} \sim \mathcal{N}(0, \sigma^2)\), which corresponds to random Fourier features \cite{tancik2020fourier}. In this case, the parameter \(\sigma\) determines the spread of frequencies and therefore the effective bandwidth of the representation. Larger values of \(\sigma\) introduce higher-frequency components, enabling the model to capture fine-scale variations, while smaller values bias the representation toward smoother functions. 
Beyond random sampling, structured choices of \(\mathbf{B}\) have also been explored. For instance, frequencies can be arranged in a deterministic, often logarithmic progression to ensure coverage across multiple scales \cite{mildenhall2021nerf}. Such multiscale constructions assign features to both low- and high-frequency bands in a controlled manner, improving stability while retaining the ability to represent fine details. This is particularly useful in high-resolution settings where uniform random sampling may lead to uneven spectral coverage. However, such structured encodings can introduce axis bias, as frequencies are typically aligned with coordinate axes rather than distributed isotropically \cite{tancik2020fourier}.  An alternative approach is to treat $\mathbf{B}$ as a learnable matrix. In this case, the model adapts its frequency basis during training, effectively discovering task-dependent spectral representations. While this increases flexibility, it can also introduce optimization challenges, as the network must jointly learn both the basis and the combination weights. Consequently, fixed or partially structured choices of $\mathbf{B}$ are often preferred for their stability and inductive bias. Such partial designs include fixing frequency directions while learning their scales, or employing multiscale encodings with predefined frequency bands and learnable coefficients, as in hash-grid representations \cite{muller2022instant}. Overall, the matrix \(\mathbf{B}\) serves as a mechanism for shaping the spectral support of the input representation. Its design determines not only which frequencies are available, but also how efficiently the network can approximate signals with varying degrees of complexity. Careful selection of \(\mathbf{B}\) is therefore crucial for balancing expressivity, stability, and generalization in implicit neural representations.

\subsubsection{\textcolor{blue}{Injecting Oscillations: Periodic Activations for Signal Representation}}
\begin{figure}[t]
    \centering
    \includegraphics[width=\linewidth]{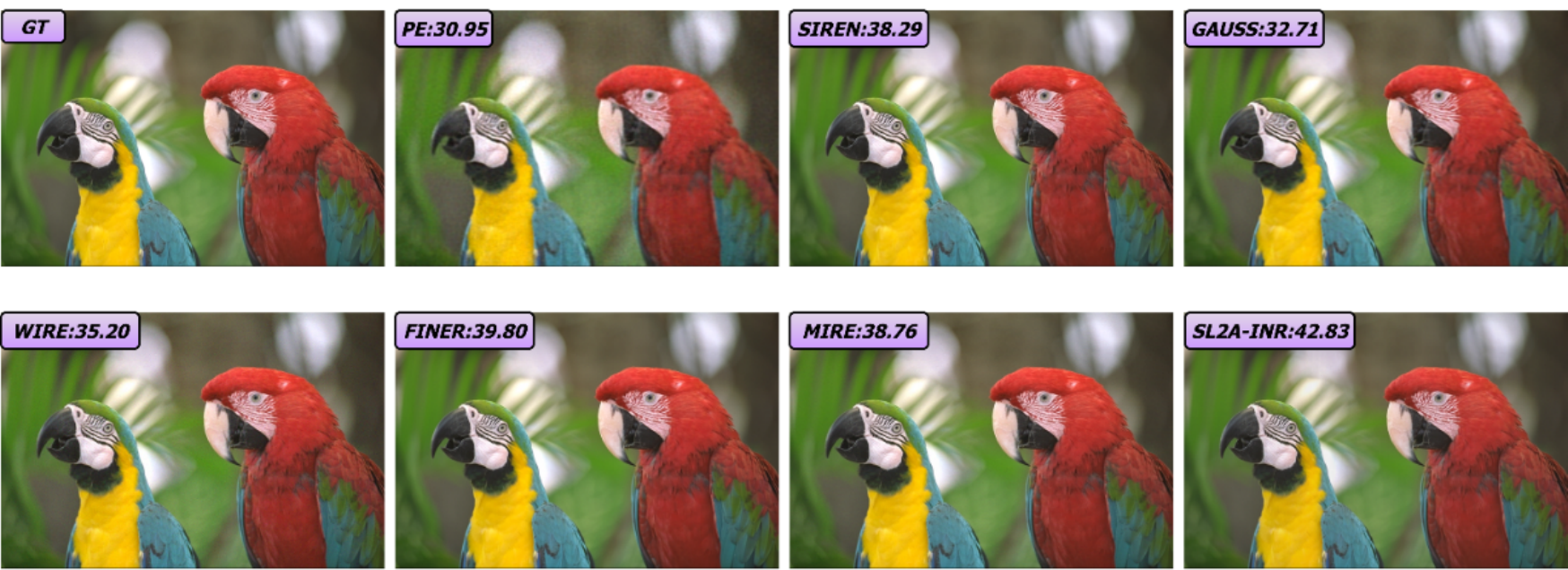}
\caption{Representation capacity of different INR formulations on a single image fitting task. Each model overfits the same image; PSNR highlights how improved spectral handling, localization, and adaptivity yield higher-fidelity reconstructions.} \label{fig:other activations}
\end{figure}

Although positional encoding (PE) significantly improves the ability of INRs to capture high-frequency components, the reconstructed signals still exhibit a bias toward low-frequency structure. As shown in \Cref{fig:other activations}, comparisons between the ground truth (GT) image and the PE-based reconstruction—particularly in regions such as the bird’s eye and feathers—reveal residual blurring of fine details. The figure further illustrates that several subsequent INR variants recover these structures more faithfully, suggesting that PE alone does not fully mitigate the low-frequency bias. The numbers next to each model denote the corresponding PSNR values, highlighting these improvements and motivating the refinements discussed in this section.

SIREN\cite{sitzmann2020implicit} addresses this limitation by replacing conventional pointwise nonlinearities with periodic activations, so that each layer takes the form \( \mathbf{h}^{(\ell+1)}=\sin\!\big(\mathbf{W}^{(\ell)}\mathbf{h}^{(\ell)}+\mathbf{b}^{(\ell)}\big) \). This seemingly simple change is important: by embedding oscillatory structure directly into the network, SIREN can represent fine signal variation much more naturally than standard ReLU-based coordinate networks. Moreover, because the derivative of a sinusoid is again a sinusoid up to a phase shift, the network is well suited not only for modeling signals themselves, but also for modeling their derivatives, which is essential in applications involving differential constraints and boundary value problems. To enable stable training, SIREN introduces a principled initialization scheme designed to preserve the distribution of activations across layers. Specifically, weights are initialized as \( W^{(\ell)}_{ij} \sim \mathcal{U}\!\big(-\sqrt{6/n_\ell},\, \sqrt{6/n_\ell}\big) \), which ensures that pre-activations are approximately normally distributed and that the outputs of the sine nonlinearity follow a stable distribution, and this prevents signal amplification or attenuation as it propagates through the network. The first layer is treated differently by introducing a frequency scaling factor, \( \mathbf{h}^{(1)} = \sin\!\big(\omega_0 (\mathbf{W}^{(0)} \mathbf{x} + \mathbf{b}^{(0)})\big) \), where \( \omega_0 = 30 \) is chosen to allow the network to represent high-frequency variation directly from the input coordinates. In addition, the authors note that training can be further accelerated by applying this frequency factor across all layers through a reparameterization of the weights. Writing \( \mathbf{W} = \tilde{\mathbf{W}} \cdot \omega_0 \) and initializing \( \tilde{W}^{(\ell)}_{ij} \sim \mathcal{U}\!\big(-\sqrt{6/(\omega_0^2 n_\ell)},\, \sqrt{6/(\omega_0^2 n_\ell)}\big) \) preserves the same activation statistics while effectively scaling gradients with respect to the weights. This accelerates optimization without altering the forward signal propagation. Overall, SIREN is not merely an INR with a different activation, but a coordinated design in which periodic nonlinearities, activation statistics, and initialization are jointly matched. This combination enables both high-frequency expressivity and stable optimization, allowing SIRENs to accurately model signals and their derivatives in applications such as boundary value problems and physics-informed learning.

\subsubsection{\textcolor{blue}{Beyond Pure Sinusoids: Expanding the Spectral Basis}}

Once sinusoidal activations showed that coordinate MLPs can recover high-frequency structure without explicit positional encoding, a deeper question emerged: is periodicity itself essential, or is it simply one mechanism for producing a suitable hidden representation? Ramasinghe et al.\cite{ramasinghe2022beyond} ~shift the focus away from periodicity alone and toward the geometry induced by the activation in hidden space. Let \(\phi\) denote the composition of the preceding \(k-1\) layers of the network, so that \(\phi(\mathbf{x})\) is the hidden representation of the input \(\mathbf{x}\) immediately before the final linear layer. For training coordinates \(\{\mathbf{x}_n\}_{n=1}^N\), the corresponding embedding matrix is defined as \(\mathbf{X} = [\,\phi(\mathbf{x}_1), \phi(\mathbf{x}_2), \dots, \phi(\mathbf{x}_N)\,] \in \mathbb{R}^{D \times N}\), where each column of \(\mathbf{X}\) is the hidden feature vector associated with one training coordinate. Their central point is that a useful activation must shape this hidden representation so that it is rich enough to capture fine detail, yet locally well behaved enough to generalize across nearby coordinates. In this view, the singular-value structure of \(\mathbf{X}\) reflects how expressive the hidden features are, while the local Lipschitz behavior determines how sharply the representation can respond to small coordinate changes. The intuition developed in \emph{Beyond Periodicity} is that two properties of the activation function \(\psi\) are fundamental. First, its slope, as quantified by \(|\psi'(x)|\), should remain sufficiently large and controllable over relevant input intervals, enabling stable gradient propagation and preventing saturation effects. Otherwise, nearby coordinates remain too tightly coupled in feature space, and the network is biased toward overly smooth reconstructions that struggle to represent edges, textures, or other fine-scale variation. Second, that slope should itself vary meaningfully across the activation domain, or equivalently \(\psi''(x)\not\approx 0\). This allows different parts of the signal to experience different levels of local sensitivity after affine transformations move coordinates into different operating regions of the nonlinearity. Put differently, the activation should not enforce a single uniform smoothness regime everywhere; instead, it should allow the network to remain smooth where the signal is smooth while becoming more sensitive where the signal changes rapidly. From this perspective, the contribution of \emph{Beyond Periodicity} is conceptual as much as architectural. The paper argues that sinusoidal activations are effective not because periodicity is uniquely privileged, but because they satisfy these geometric requirements. This broadens the activation design space considerably. Ramasinghe et al.~show that several non-periodic activations, including Gaussian, Quadratic, Multi-Quadratic, Laplacian, and Super-Gaussian forms, can achieve similar behavior, and they further report that Gaussian activations are substantially more robust to initialization in natural-image fitting tasks. The key insight of \emph{Beyond Periodicity} is not simply that non-periodic activations may also succeed, but that the effectiveness of an activation is better understood through the geometry it induces in the hidden representation than through periodicity alone. Once this perspective is adopted, a natural next question arises: if periodicity is not the defining ingredient, what additional structure should a good activation provide?

\subsubsection{\textcolor{blue}{Localized Oscillations: Joint Space–Frequency Modeling}}

One useful perspective on this issue comes from the nature of visual signals themselves. While they contain rich high-frequency content, they are also highly localized, edges, corners, and fine textures occur at specific spatial locations rather than being uniformly distributed. Representations built from globally oscillatory components can capture frequency effectively, but they do not explicitly encode where that frequency is concentrated. This limitation is well known in classical harmonic analysis: Fourier bases provide precise frequency information but are globally supported, causing localized features to spread their influence across the entire domain. For natural images and related signals, such non-local representations are often suboptimal.

This observation motivates a refinement of INR design: the hidden representation should jointly capture oscillatory expressivity and spatial localization. In other words, the basis functions should encode not only what frequencies are present, but also where they occur in the signal. This is precisely the principle behind wavelet representations, whose atoms are localized in both space and frequency. Building on this insight, Saragadam et al.~introduced \emph{WIRE: Wavelet Implicit Neural Representations} \cite{saragadam2023wire}, where the activation is chosen as a complex Gabor wavelet rather than a pure sinusoid or pure Gaussian. For a scalar input \(x\), the WIRE activation can be  written as
\(\psi(x;\omega_0,s_0)
=
e^{j\omega_0 x}\,e^{-|s_0 x|^2}\),
where \(\omega_0\) controls the oscillation frequency and \(s_0\) controls the spatial spread of the Gaussian window. The structure of above activation is revealing. The complex exponential \(e^{j\omega_0 x}\) retains the oscillatory behavior that makes sinusoidal activations effective for representing fine detail, while the Gaussian envelope \(e^{-|s_0 x|^2}\) localizes that oscillation in space. As a result, each hidden unit behaves like a wavelet atom: it is selective to frequency, but its response is also spatially concentrated.

This shift replaces globally supported harmonics with localized oscillatory components, enabling more precise modeling of spatially varying structure.  From a visual perspective, this is highly advantageous, as localized features such as edges and fine structures can be represented without spreading error or oscillatory artifacts across the entire domain. The empirical results in WIRE \cite{saragadam2023wire} reflect exactly this behavior, showing that the method combines the frequency expressivity of sinusoidal INRs with the spatial compactness associated with Gaussian responses. Seen in this progression, WIRE is not a disconnected architectural variant, but a natural next step after \emph{Beyond Periodicity} \cite{ramasinghe2022beyond}. Once periodicity is no longer viewed as fundamental, the design space opens to activations whose advantage lies in how they localize and organize hidden features. Wavelet-inspired activations make this idea explicit: they preserve oscillatory richness, but shape it into atoms that are better matched to the spatially varying structure of natural signals.

\subsubsection{\textcolor{blue}{Learning Where to Oscillate: Adaptive Frequency Modeling}}
Yet another limitation remains. Even when the activation is well chosen, the range of frequencies that the network can effectively exploit is often still governed by fixed design choices. In sinusoidal INRs, for example, the frequency scale is largely determined by parameters such as \(\omega_0\), while in other constructions it is similarly tied to preselected activation behavior. This means that although the network may be expressive, the set of frequencies it accesses is still to a large extent prescribed in advance. Such a design can be restrictive when the signal contains a mixture of structures at very different scales. This is precisely the issue addressed by \emph{FINER: Flexible spectral-bias tuning in Implicit Neural Representation} \cite{liu2024finer}. The key observation is simple but important: in many existing INRs, the activation function has an infinite domain, yet during training the network tends to use only a relatively narrow region of that domain \cite{liu2024finer}, typically near the origin. As a result, only a limited range of oscillatory behaviors is actually activated, even if the nonlinearity itself is capable of more. FINER turns this observation into a design principle,  instead of using an activation with a fixed periodic behavior, it introduces a \emph{variable-periodic} activation whose local oscillation rate changes across its domain. Concretely, FINER uses an activation of the form
\(
\psi(x)=\sin\!\big(\omega(|x|+1)x\big).
\)
Unlike the standard sine function, whose oscillation rate is fixed everywhere, this function oscillates more rapidly as \(|x|\) increases. The consequence is that different regions of the activation correspond to different effective frequencies. Therefore, the question of frequency support is no longer controlled solely by a fixed scalar such as \(\omega_0\); it is also influenced by which part of the activation the network actually operates in. The mechanism used by FINER to access these different regimes is the bias initialization. By initializing the bias vector over a wider interval, the pre-activations \(\mathbf{W}\mathbf{x}+\mathbf{b}\) are shifted into different parts of the variable-periodic function, thereby exposing the network to a broader collection of oscillatory sub-functions. In the language of the paper, this enlarges the supported frequency set \cite{liu2024finer}. The usual SIREN-like behavior is then recovered as only a restricted portion of this broader set, whereas larger bias ranges activate additional higher-frequency regimes. This gives FINER a very different flavor from earlier INR designs. The goal is not merely to choose a better activation, nor only to localize the atoms more effectively, but to make the spectral bias itself more flexible. From the neural tangent kernel viewpoint \cite{liu2024finer,tancik2020fourier}, the paper further shows that enlarging the bias range strengthens the kernel’s diagonal structure and increases the number of large eigenvalues, both of which favor faster learning of high-frequency components. 

\subsubsection{\textcolor{blue}{Shaping Spectra Through Weight Reparameterization}}
A natural next question is whether this adaptation must come entirely from the activation function. Methods such as SIREN, WIRE, and FINER all improve INR performance by reshaping the functional building blocks themselves, through periodicity, localization, or variable oscillation. Yet from a signal-processing perspective, representation quality depends not only on the basis-like atoms available to the model, but also on how the coefficients of those atoms are learned during optimization. This motivates a different line of thought, instead of modifying the input encoding or the activation, can one alter the \emph{training parameterization} of the network so that high-frequency components are learned more effectively? This is the idea behind \emph{Fourier Reparameterized Training} \cite{shi2024improved}. Rather than learning each weight matrix \(\mathbf{W}^{(\ell)}\) directly, the method writes it as
\(
\mathbf{W}^{(\ell)}=\mathbf{\Lambda}^{(\ell)}\mathbf{B}^{(\ell)},
\)
where \(\mathbf{\Lambda}^{(\ell)}\) is a trainable coefficient matrix and \(\mathbf{B}^{(\ell)}\) is a fixed matrix whose rows are sampled Fourier bases. In this form, the network does not optimize arbitrary weights in the usual coordinate system; instead, it learns how to combine a predefined collection of spectral components in order to form the effective linear transformation. The interpretation is quite natural. Classical signal representations often separate a signal into a basis and corresponding coefficients. Fourier reparameterization imports this idea into INR training: the fixed matrix \(\mathbf{B}^{(\ell)}\) provides a spectral organization of the parameter space, while the trainable coefficients \(\mathbf{\Lambda}^{(\ell)}\) determine how strongly different components participate in the learned mapping. The important point is that this changes the \emph{optimization geometry} rather than the final inference architecture. After training, the product \(\mathbf{\Lambda}^{(\ell)}\mathbf{B}^{(\ell)}\) is simply collapsed back into an ordinary weight matrix, so the representation at inference remains an MLP of the same form. This is especially important for INRs because spectral bias reflects not only representational limitations, but also the dynamics of optimization. The paper shows that an appropriate reparameterization can mitigate the imbalance between low- and high-frequency gradient contributions, thereby improving the network’s ability to recover fine-scale detail.

\subsubsection{\textcolor{blue}{Signal-Matched Activations: Choosing Bases from Data}}
The developments so far have progressively expanded the design space of INRs, through improved activations, better localization, adaptive frequency support, and more favorable training parameterizations. Yet, a fundamental limitation still remains: in all these approaches, the structure of the representation is largely fixed \emph{a priori}. The same activation function, or a small family of parameterized variants, is used uniformly across layers, regardless of the specific signal being represented. This stands in contrast to classical signal processing, where representations are often tailored to the signal itself, for example through adaptive bases or dictionary learning. This observation motivates a further step, rather than designing a single “best” activation, can one construct an INR whose representation is \emph{matched} to the signal? This is precisely the idea behind \emph{Matched Implicit Neural Representations (MIRE)} \cite{jayasundara2025mire}.  MIRE departs from the standard paradigm by allowing different layers of the network to use different activation functions, selected adaptively based on how well they align with the signal. Formally, instead of fixing a single nonlinearity \(\sigma\) across all layers, MIRE considers a dictionary of activation atoms \(\{\psi_k\}\), drawn from functions with diverse space–frequency characteristics, including sinc \cite{saratchandran2024sampling}, raised cosine (RC) \cite{jayasundara2025mire, thennakoon2025cosmo}, Gaussian \cite{ramasinghe2022beyond}, Gabor wavelets \cite{saragadam2023wire}, prolate spheroidal wave-functions \cite{jayasundara2025pin}, and sinusoids \cite{sitzmann2020implicit}. Each layer of the network is then assigned an activation chosen from this dictionary, resulting in a representation of the form
\(
\mathbf{h}^{(\ell+1)} = \psi_{k_\ell}\!\big(\mathbf{W}^{(\ell)}\mathbf{h}^{(\ell)}+\mathbf{b}^{(\ell)}\big),
\)
where the index \(k_\ell\) is selected to best match the signal at layer \(\ell\). The key mechanism used to determine this sequence is inspired by dictionary learning and matching pursuit. MIRE proceeds layer by layer, selecting at each stage the activation that minimizes the reconstruction error, while keeping previously selected activations fixed. In this way, the network constructs a sequence of nonlinearities that is specifically adapted to the signal being represented, rather than relying on a globally fixed design. This can be interpreted as moving from a fixed basis representation to an adaptive one. Earlier INR designs implicitly assume a predetermined set of atoms, whether global sinusoids, localized wavelets, or variable-periodic functions. MIRE, in contrast, learns a \emph{composition of atoms} that is tuned to the signal, much like selecting basis functions in a learned dictionary. This allows the representation to capture different regimes of the signal, smooth regions, sharp transitions, and localized structures, using different functional components across layers.

\subsubsection{\textcolor{blue}{From Sampling Theory to Neural Representations}}

A complementary viewpoint emerges when one asks what class of signals an INR can reconstruct from sampled observations. Rather than viewing an INR solely as a generic function approximator, Saratchandran \emph{et al.} \cite{saratchandran2024sampling} show that, under suitable conditions, its activation can be interpreted as a generator whose translated copies span a signal space. This links INR reconstruction to classical sampling theory, where an $\Omega$-bandlimited signal $f$ can be reconstructed from its samples through the Shannon expansion
$
f(x)=\sum_{n=-\infty}^{\infty} f\!\left(\frac{n}{2\Omega}\right)\,\mathrm{sinc}\!\left(2\Omega x-n\right).
$
Motivated by this analogy, they consider a generator function $F$, coefficients $a(k)\in \ell^2(\mathbb{R})$, and the associated shift-invariant signal space
$
V(F)=\left\{s(x)=\sum_{k\in\mathbb Z} a(k)\,F(x-k): a\in \ell^2(\mathbb{R})\right\},
$
where $x\in\mathbb{R}$ denotes the continuous coordinate and $k\in\mathbb Z$ indexes translated copies of $F$. The central question is then whether the family of translates $\{F(x-k)\}_{k\in\mathbb Z}$ provides a stable system for reconstruction. In their framework, this is formalized through the notions of a \emph{Riesz basis} and a \emph{weak Riesz basis}: the former guarantees stable and unique reconstruction within $V(F)$ together with the partition-of-unity condition, while the latter satisfies only the stability requirement. Under this formulation, a two-layer INR with activation $F$ can approximate any signal $g\in V(F)$ whenever $\{F(x-k)\}_{k\in\mathbb Z}$ forms a weak Riesz basis. To move beyond $V(F)$ and approximate arbitrary signals in $L^2(\mathbb{R})$, however, the stronger partition-of-unity condition becomes essential through the corresponding scaled spaces $V_\Omega(F)$ \cite{saratchandran2024sampling}. This yields a principled comparison of activations: sinc generates a Riesz basis, Gaussian activations generate only a weak Riesz basis, wavelets in general also yield weak Riesz bases, whereas fixed-frequency sinusoidal and ReLU activations do not satisfy the required basis conditions. Their main conclusion is therefore that sinc activation is optimal in this reconstruction-theoretic sense, among the activations they analyze, it is the one that supports stable reconstruction within the associated space and, through scaling, enables approximation of arbitrary $L^2(\mathbb{R})$ signals \cite{saratchandran2024sampling}. 


\subsubsection{\textcolor{blue}{Learning the Activation Itself: Toward Adaptive Basis Functions}}
This sampling-theoretic perspective highlights a key design question, the choice of the generator function $F$ fundamentally determines the reconstruction space and its stability properties. While the analysis in \cite{saratchandran2024sampling} shows that sinc activations are optimal in the sense that their translates form a Riesz basis and enable stable approximation of arbitrary $L^2(\mathbb{R})$ signals, this optimality is tied to a fixed generator. In practice; however, natural signals are rarely globally bandlimited and often exhibit spatially varying spectral characteristics, making a single fixed activation suboptimal.
Related in spirit, recent works have also explored adaptive activation design. In particular, SL2A-INR \cite{rezaeian2025sl2a} parameterizes the activation itself as a learnable function
$
\psi(x) = \sum_{k=0}^{K} a_k\, T_k(\sigma(x)),
$
where $T_k$ are Chebyshev polynomials and the coefficients $a_k$ are learned during training. As higher-degree polynomial terms capture increasingly high-frequency components, learning their coefficients allows the network to flexibly adjust its spectral bias. Instead of relying on a predetermined generator, the model effectively learns the basis in which the signal is represented. Importantly, this learnable activation is placed at the front of the network and its output modulates subsequent layers, ensuring that the learned spectral structure influences the entire representation. This changes the role of the activation, it is no longer a fixed nonlinearity, but a data-adaptive mechanism that redistributes representational capacity across frequencies.

\begin{figure*}[t]
    \centering
    \includegraphics[width=\linewidth]{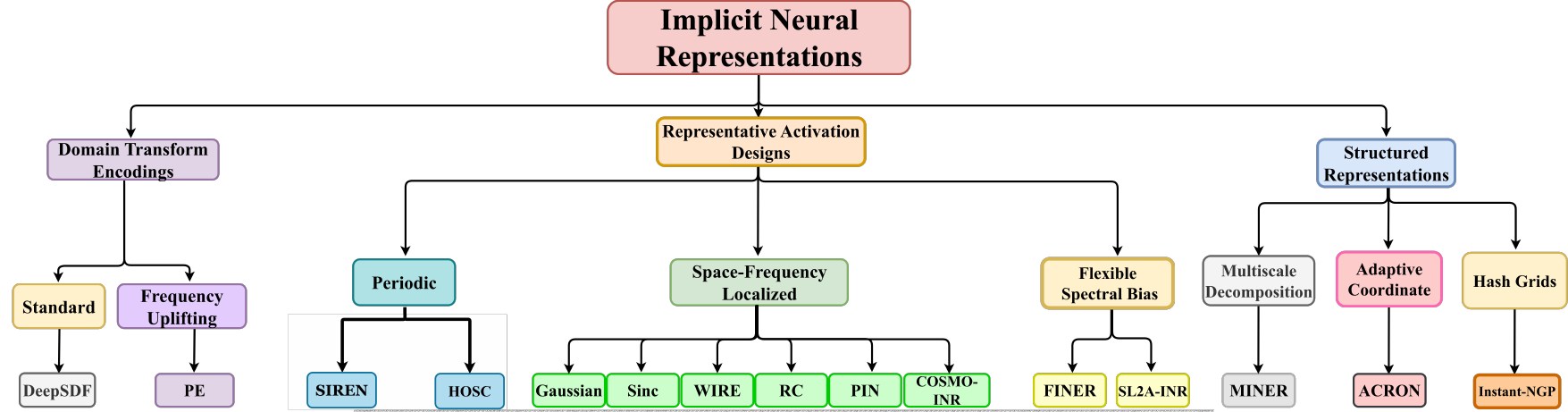}
\caption{Hierarchical classification of INR methods based on architectural design and signal support.}\label{fig:tree_diagram}
\end{figure*}


\subsubsection{\textcolor{blue}{Further Approaches to Spectral Design}}A particularly insightful alternative perspective is provided by multiplicative filter networks (MFNs) \cite{fathony2020multiplicative}, which move away from standard compositional architectures and instead build representations through products of simple input-dependent filters. By combining sinusoidal or Gabor-type responses multiplicatively, MFNs can be interpreted as implicitly spanning a large family of Fourier- or Gabor-like basis functions, making the role of spectral atoms more explicit. In this sense, they offer a useful bridge between INR design and classical transform-based thinking, rather than relying only on depth to gradually synthesize oscillatory structure, they generate complex patterns through interactions among simpler harmonic components. A related direction recognizes that spectral design need not remain globally fixed, but can instead be conditioned on prior knowledge about the signal. INCODE \cite{kazerouni2024incode} embodies this idea through a harmonizer–composer framework, where embeddings obtained from a task-specific pretrained model dynamically modulate key activation parameters of the INR, allowing the representation to adapt more effectively to the underlying signal structure. This replaces a static harmonic dictionary with a conditioned one, whose effective frequency response adapts to the signal class or reconstruction task. More broadly, INCODE suggests that spectral design in INRs can be driven not only by architectural priors, but also by external semantic or task-level information, bringing it closer to adaptive transform design and data-dependent dictionary learning. More recent works further broaden this design space by moving beyond fixed sinusoidal parameterizations toward activations with stronger localization and better control over spectral propagation. A representative example is PIN \cite{jayasundara2025pin}, which introduces prolate spheroidal wave functions (PSWFs) as INR activations. As PSWFs are optimally concentrated in both space and frequency, they provide oscillatory atoms that remain localized, making them attractive for representing fine detail without introducing excessive artifacts in smooth regions. Closely related in spirit, HOSC \cite{serrano2024hosc} introduces a periodic activation with a controllable sharpness parameter, allowing the nonlinearity to transition from smooth oscillatory behavior toward sharper waveforms so that abrupt transitions and high-frequency features can be preserved more effectively while retaining differentiability and compactness \cite{serrano2024hosc}. A related but distinct direction asks not only what spectral bias an activation introduces at a single layer, but also how that spectral content evolves as signals propagate through depth. COSMO-INR\cite{thennakoon2025cosmo} follows this perspective by studying how complex sinusoidal modulation of the activation can enrich spectral support across the network. Using harmonic analysis and Chebyshev-polynomial arguments, it shows that common activations can attenuate portions of the spectrum after each nonlinear transformation, and proposes complex sinusoidal modulation to preserve richer spectral support throughout the network. Taken together, these developments indicate that spectral design in INRs is evolving from fixed harmonic constructions toward more explicit, localized, adaptive, and propagation-aware continuous dictionaries.  
A visual summary of the discussed activation functions and their corresponding functional forms is provided in \Cref{tab:inr_activations_visual}, and the broader taxonomy of INR models, highlighting their major representation mechanisms, is shown in \Cref{fig:tree_diagram}.
\subsubsection{\textcolor{blue}{Addressing the Limitations of Fixed Spectral Parameterizations}}A recurring practical challenge in INR design is that the spectral behavior of the representation is often governed by a small number of hyperparameters whose optimal values depend strongly on the target signal. For instance, in SIREN, this role is played by the frequency scale \(\omega_0\); in WIRE, by the wavelet parameters \(\omega_0\) and \(s_0\), which control oscillation and localization; and in Fourier-feature-based models, by the scale of the coordinate embedding. Despite these parameters strongly affect performance, they are usually fixed to default values or tuned by costly grid search \cite{saragadam2023wire,saratchandran2024sampling}, even though the best setting can vary substantially across signals. The difficulty is that these hyperparameters effectively determine the frequency support that the network can access most easily during training. If this support is poorly matched to the spectrum of the target signal, the INR may still possess nominal high-frequency capacity, yet fail to exploit it efficiently. In practice, this mismatch often appears as slow convergence, blurred structure, or insufficient recovery of fine detail, even in architectures specifically designed to mitigate spectral bias. FreSh~\cite{kania2024fresh} addresses this issue by shifting attention away from architecture design itself and toward the alignment between the signal spectrum and the spectrum of the model at initialization. Its central observation is that the output of an untrained INR already reveals the frequencies that the model is predisposed to represent well. If that initial spectrum is close to the spectrum of the target signal, the subsequent optimization is more effective; if not, training must first overcome this spectral mismatch. Based on this idea, FreSh selects hyperparameters by comparing the target signal \(A\) and the model output at initialization through their spectra. The signal spectrum is summarized by a vector \(S(A)\) derived from the discrete Fourier transform \cite{smith1997scientist}, and the chosen configuration is the one whose initialized output has the closest normalized spectrum, measured using the Wasserstein distance. Thus, instead of training many candidate models and selecting the best one afterward, FreSh chooses the configuration whose frequency distribution is already most compatible with the signal before learning begins.

\subsection{Structured Representations in Implicit Neural Fields}
A different but closely related way to understand INRs is to move away from the viewpoint of individual activations or embeddings, and instead examine how the representation itself is constructed. A key question that naturally arises is: \emph{what are the elementary components that an INR uses to synthesize a signal, and how are these components organized?}

A more precise answer to this question is provided by Y{\"u}ce et al.~\cite{yuce2022structured}. Rather than treating the network as a black-box function approximator, they show that a broad class of INR architectures can be interpreted through a structured dictionary viewpoint, where the represented signal is formed from harmonic components induced by the input mapping and the nonlinearities. The key mechanism can be illustrated with a simple example. Consider an input mapping of the form \(\gamma(x) = e^{j\omega x}\), and suppose the activation admits a polynomial expansion of the form \(\rho(z) = \sum_{k=0}^{K} \alpha_k z^k\). This assumption is not restrictive, as most analytic activation functions can be well approximated by low-order polynomials (e.g., via Taylor or Chebyshev expansions), while even non-smooth activations such as ReLU can be approximated in a similar manner. The polynomial form is particularly convenient for analysis, as it allows the spectral effect of the activation to be characterized explicitly through interactions of frequency components. Passing \( \gamma(x) \) through the nonlinearity then gives \( \rho(\gamma(x)) = \sum_{k=0}^{K} \alpha_k e^{jk\omega x} \), which is a superposition of integer harmonics of the original frequency. This is the central insight emphasized in the paper: once nonlinear activations are applied to an initial frequency mapping, they do not merely preserve those frequencies, but generate new harmonics, and repeated composition across layers progressively enriches the available spectrum. Y{\"u}ce et al. therefore interpret INRs of this form as representing signals through structured harmonic combinations derived from the frequencies present in the input mapping. From this perspective, the network acts as a structured signal dictionary, whose atoms are shaped by the initial mapping frequencies and by the harmonic interactions induced through depth. This viewpoint clarifies both the expressive power of INRs and the fact that their representable frequency content remains fundamentally linked to the choice of input mapping. While this perspective emphasizes how INRs construct signals through harmonic expansions, an alternative line of work focuses on how signal content is organized across scales. In classical signal processing, multiscale decompositions provide a complementary view, where a signal is not represented all at once, but progressively refined from coarse structure to fine detail. MINER builds directly on this principle by adopting a Laplacian pyramid representation~\cite{saragadam2022miner}. Instead of modeling the full signal in a single step, the signal is first approximated at a very low resolution, capturing its dominant, low-frequency structure. Higher-resolution levels then model only the residual information needed to refine this coarse approximation. In this sense, each scale contributes the “missing details” that were not explained at coarser levels. This hierarchical refinement has two important implications. First, it naturally separates information across scales, with coarse levels capturing global structure and finer levels focusing on localized high-frequency content.

\clearpage
\newcolumntype{C}[1]{>{\centering\arraybackslash}m{#1}}
\newcolumntype{Y}[1]{>{\raggedright\arraybackslash}m{#1}}

\begin{table*}[p]
\centering
\footnotesize
\caption{Qualitative summary of representative activation functions used in INRs. Here, $\omega$, $\omega_1$, $\omega_2$, and $s$ denote activation parameters controlling oscillation and localization.}
\label{tab:inr_activations_visual}
\renewcommand{\arraystretch}{1.1}
\setlength{\tabcolsep}{4pt}

\begin{tabular}{C{1.8cm} C{3.1cm} C{2.3cm} @{\hspace{0.65cm}} Y{5.5cm}}
\hline
\textbf{Representative design} & \textbf{Form} & \textbf{Plot} & \textbf{Qualitative behavior} \\
\hline

ReLU \cite{tancik2020fourier}
& $\max(0,x)$
& \includegraphics[height=1.8cm]{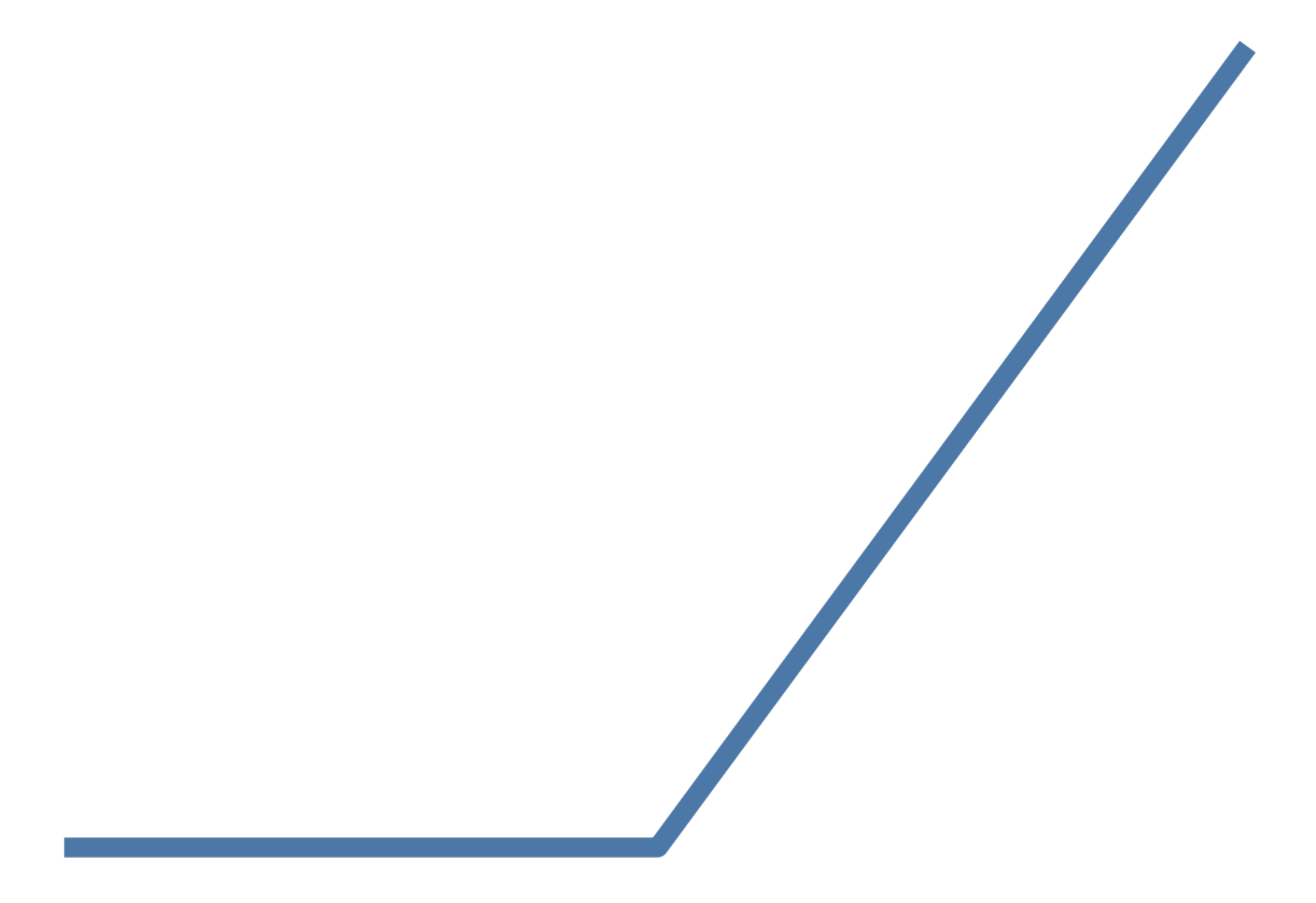}
& Piecewise linear; non-smooth. Low-frequency bias; limited detail recovery. \\
\hline

PE \cite{tancik2020fourier}
& $\gamma(x)=[\sin(\omega_k x),\cos(\omega_k x)]_{k=1}^{K}$
& \includegraphics[height=1.8cm]{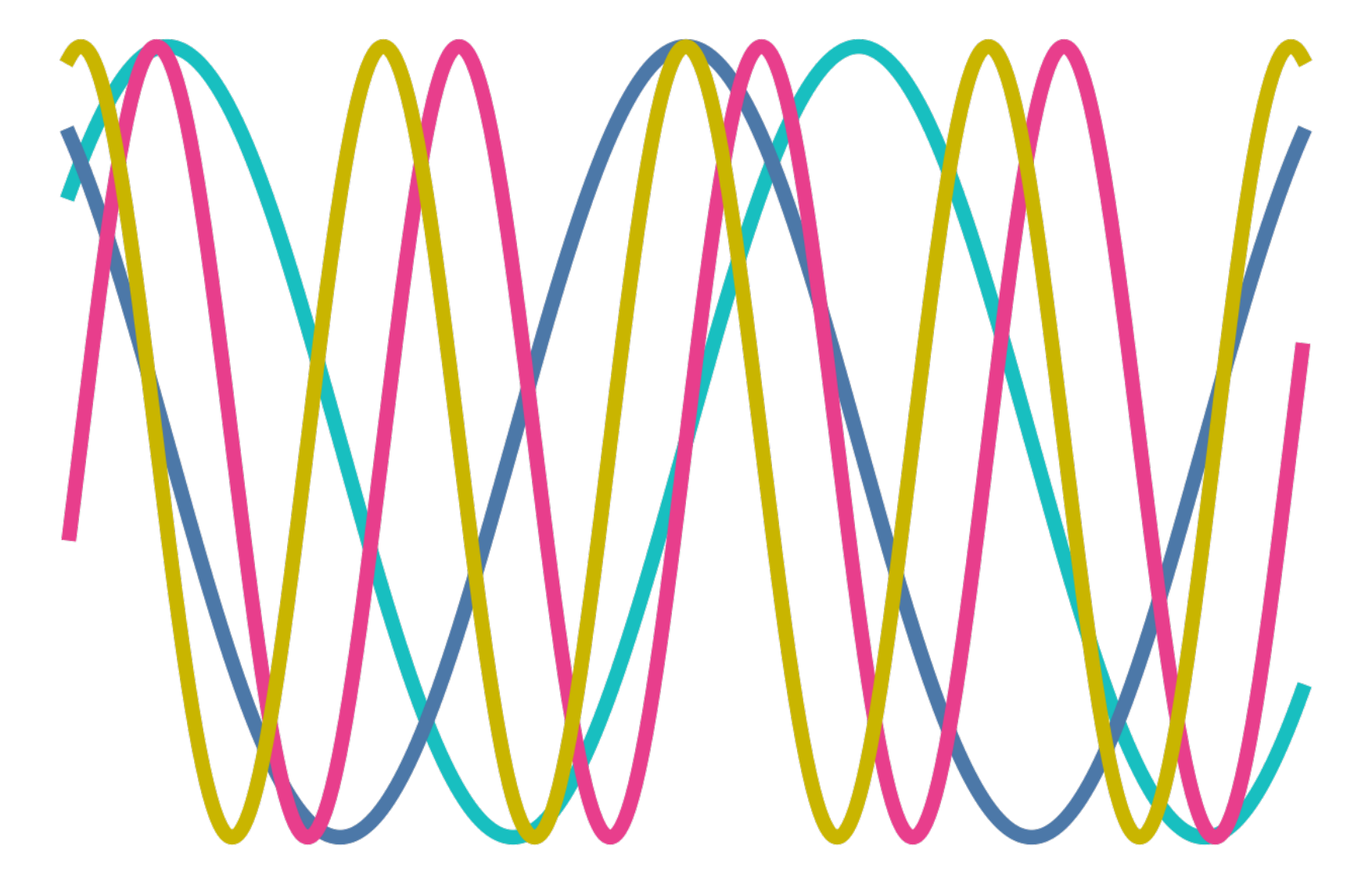}
& Multi-frequency lifting through projected sinusoidal channels; improves high-frequency representation before the MLP. \\
\hline

SIREN \cite{sitzmann2020implicit}
& $\sin(\omega x)$
& \includegraphics[height=1.8cm]{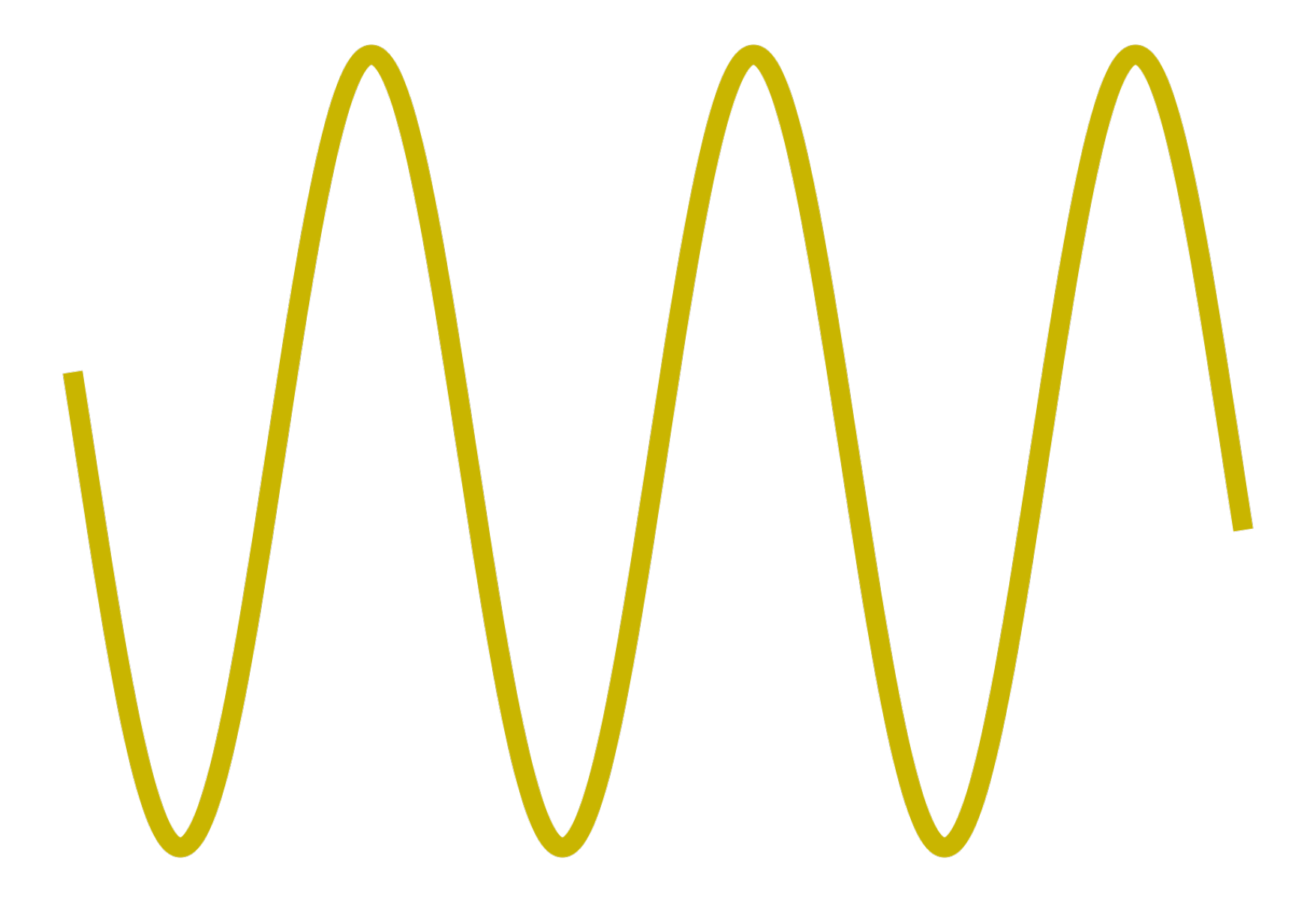}
& Globally oscillatory; rich harmonics. Accurate derivatives and fine detail. \\
\hline

Gaussian \cite{ramasinghe2022beyond}
& $\exp(-|s|x^2)$
& \includegraphics[height=1.8cm]{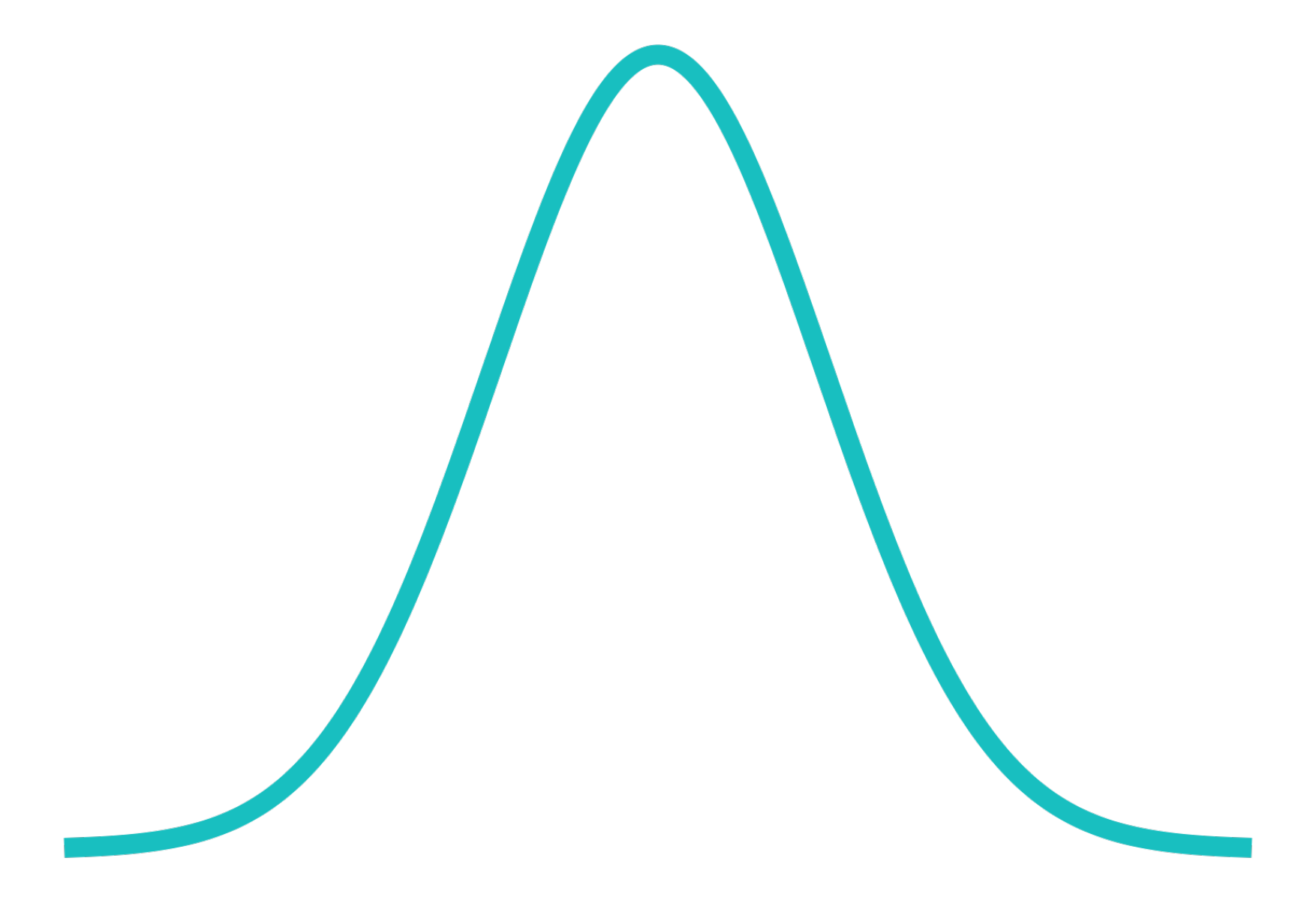}
& Smooth and localized; stable fitting of local structures. \\
\hline

WIRE \cite{saragadam2023wire}
& $\exp(-|s|x^2)\cos(\omega x)$
& \includegraphics[height=1.8cm]{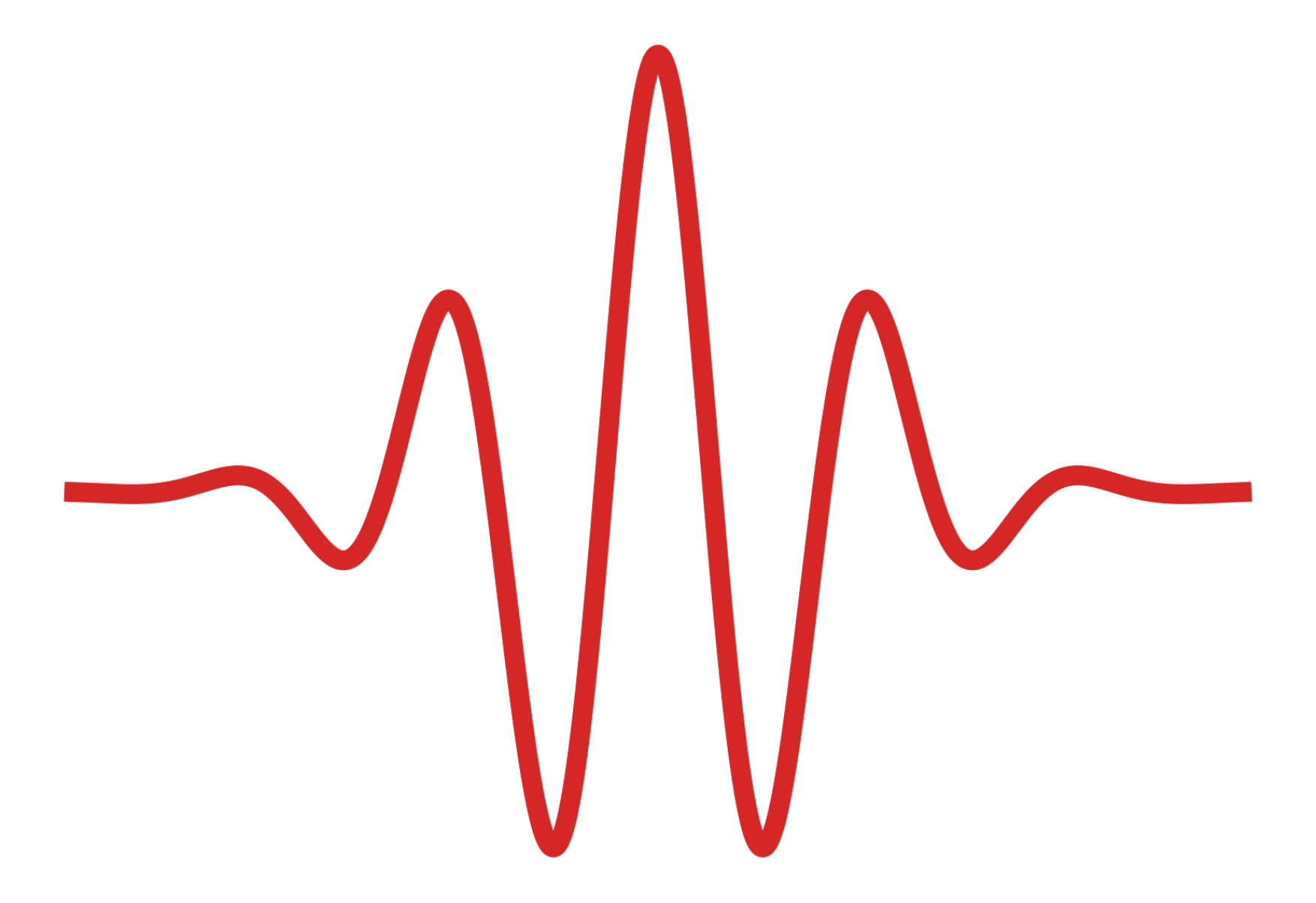}
& Localized oscillation; joint spatial--frequency concentration. \\
\hline

FINER \cite{liu2024finer}
& $\sin\!\big(\omega(|x|+1)x\big)$
& \includegraphics[height=1.8cm]{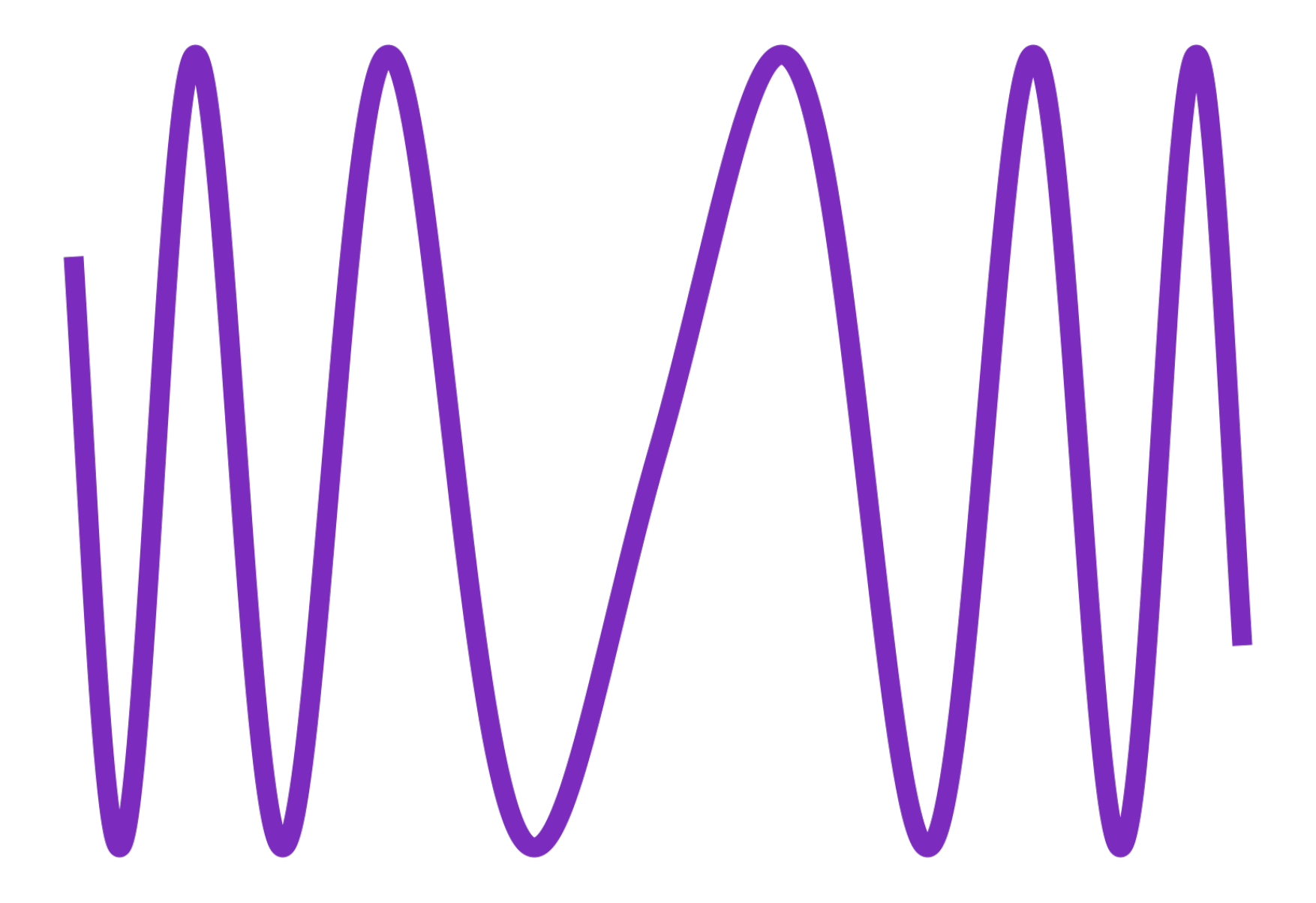}
& Input-adaptive oscillation; flexible spectral support. \\
\hline

HOSC \cite{serrano2024hosc}
& $\tanh(\beta\sin(\omega x))$
& \includegraphics[height=1.8cm]{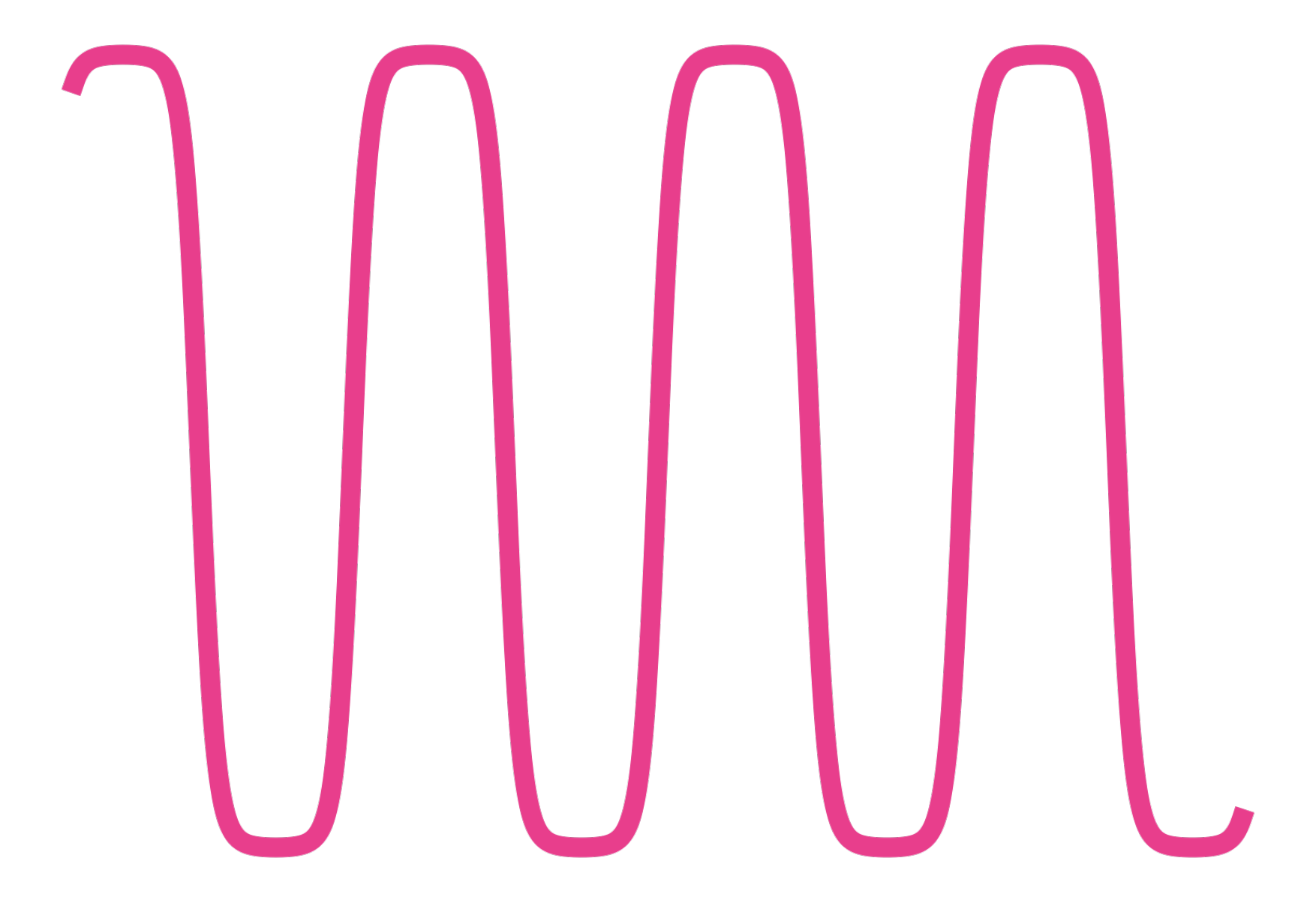}
& Controlled gradients with oscillation; improved multiscale stability. \\
\hline

RC \cite{jayasundara2025mire, thennakoon2025cosmo}
& $\dfrac{\operatorname{sinc}(\omega_1 x)\cos(\omega_2 x)}{1-|s|x^2}$
& \includegraphics[height=1.8cm]{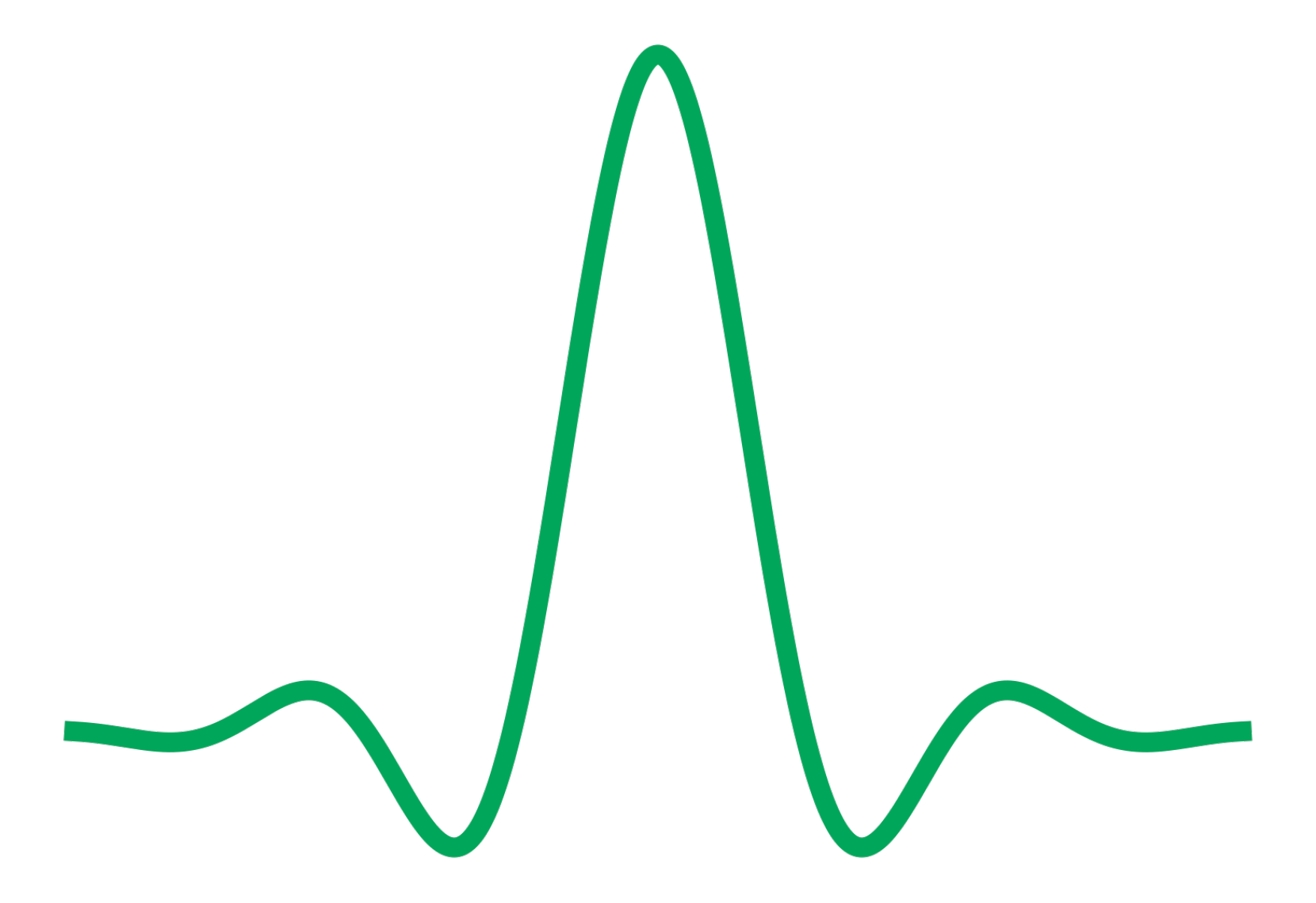}
& Localized with controlled bandwidth, and eliminates frequency blind spots by breaking the activation symmetry.  \\
\hline

PIN \cite{jayasundara2025pin}
& Eigenfunctions of the time--band limiting concentration operator
& \includegraphics[height=1.8cm]{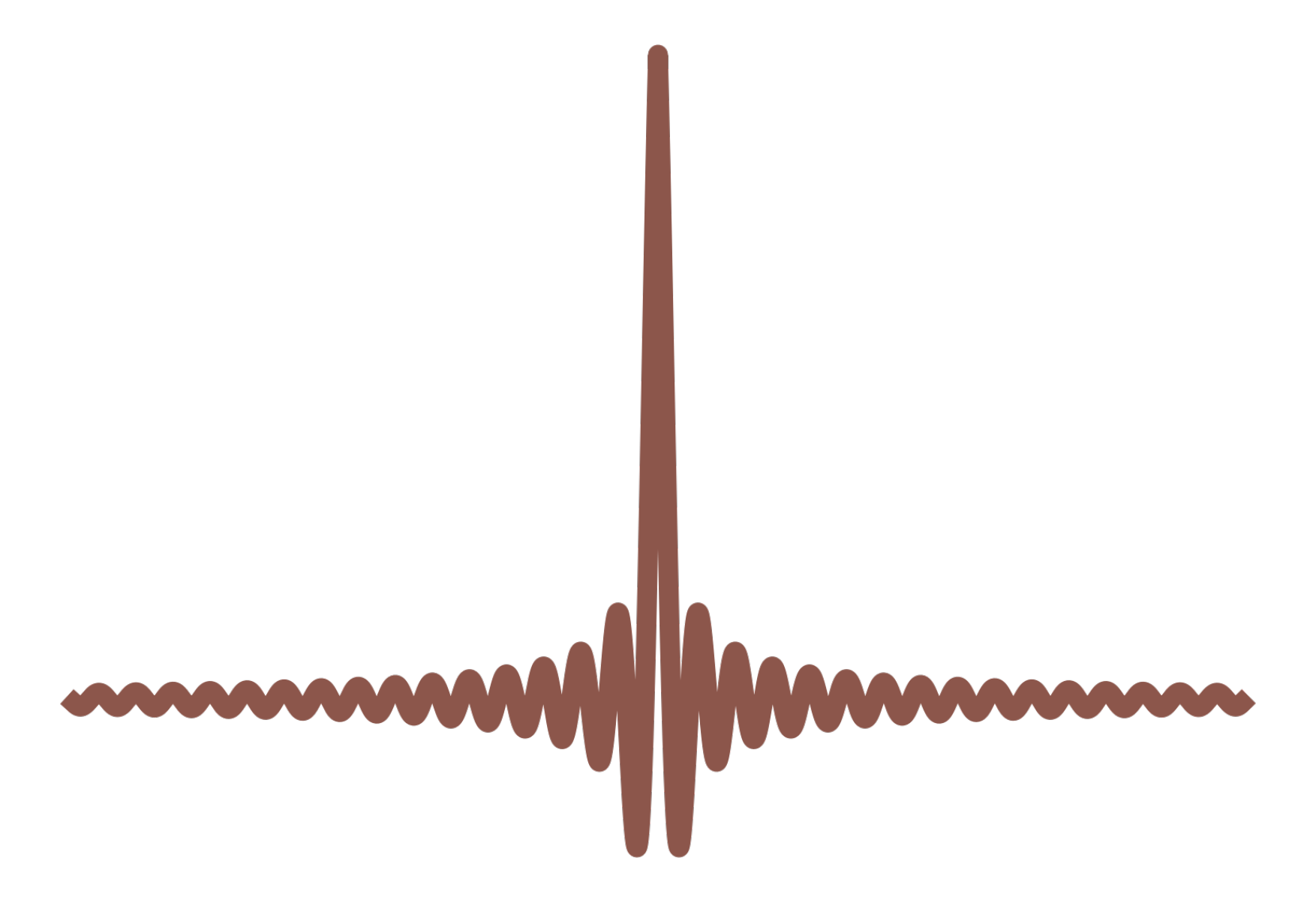}
&  Strong space--frequency concentration for compact and structured representations. \\
\hline

Sinc \cite{saratchandran2024sampling}
& $\operatorname{sinc}(\omega x)$
& \includegraphics[height=1.8cm]{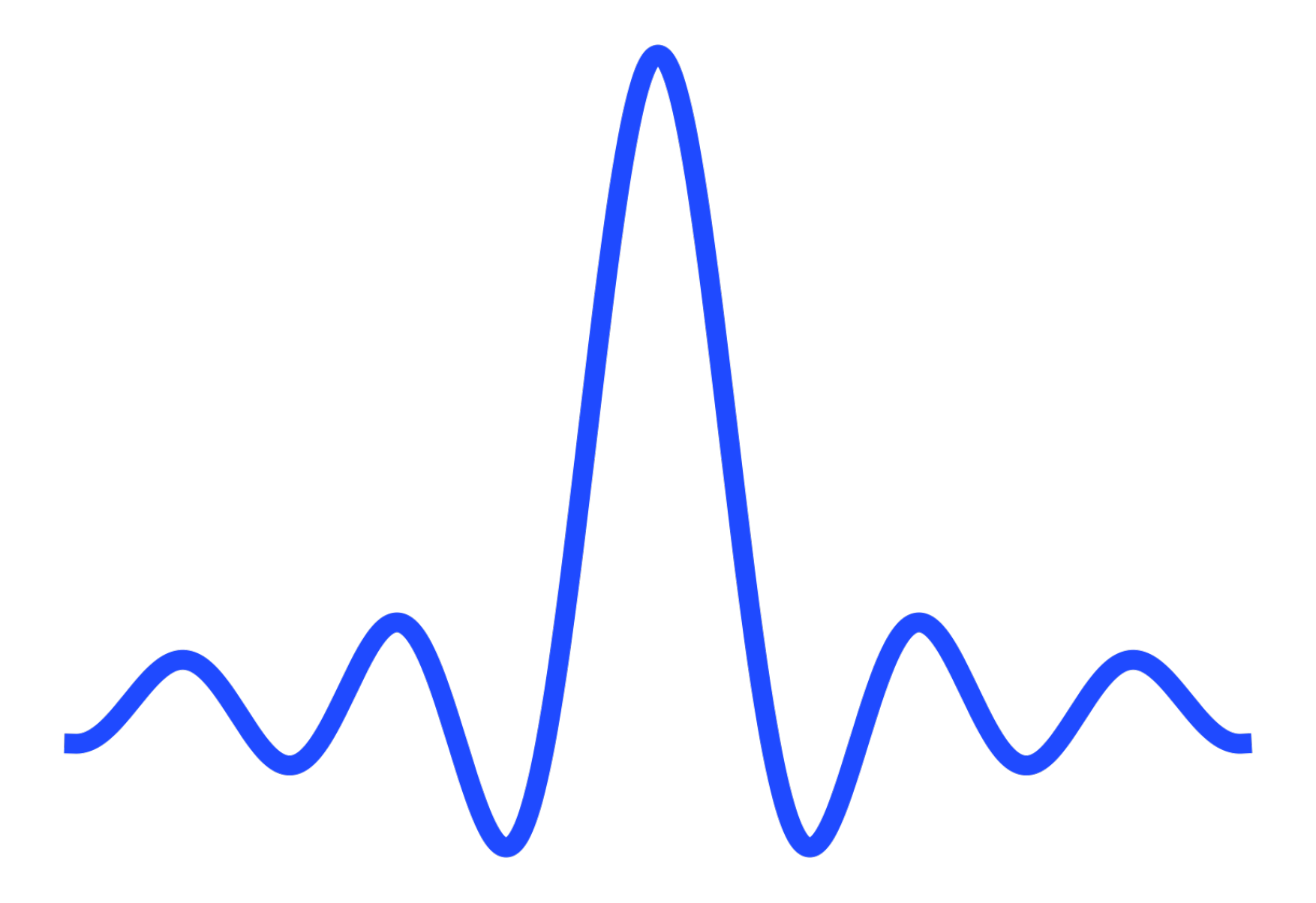}
& Band-limited interpolation kernel, and provides superior high-frequency reconstruction \\
\hline

\end{tabular}
\end{table*}

\clearpage

  Because the Laplacian decomposition produces components that are approximately orthogonal, each level adds largely new information with limited redundancy. Second, the residuals at finer scales are often sparse: many regions contain little additional detail once the coarse structure is known. MINER leverages this structure by replacing a single global network with a collection of small, local MLPs assigned to disjoint spatial patches at each scale. Each network is responsible only for modeling the residual content within its region, rather than the entire signal. Training proceeds in a coarse-to-fine manner, so that the model first captures the dominant structure and then incrementally refines it with higher-frequency details. An important consequence of this formulation is efficiency. Since many regions have negligible residual energy at finer scales, the corresponding local models can be skipped altogether. This adaptive allocation of capacity, combined with the multiscale decomposition, allows MINER to represent large signals accurately while using significantly fewer parameters and computation.

While MINER decomposes the signal across scales, ACORN \cite{martel2021acorn} introduces spatial adaptivity through an explicit multiscale block decomposition of the input domain. In two dimensions, ACORN uses a quadtree, and in three dimensions, an octree, with each spatial location assigned to a single \emph{active} block at one scale rather than being represented simultaneously across multiple resolutions. The set of active blocks is optimized online during training, so that regions with higher fitting error are allocated finer blocks, while smoother or empty regions remain at coarser scales or are pruned altogether. Architecturally, ACORN is a hybrid implicit-explicit representation. A coordinate encoder maps the global block index, which encodes both position and scale, to a local feature grid, and a lightweight decoder then predicts the signal value from features obtained by interpolation at continuous local coordinates within the block. This design allows the expensive computation to be amortized at the block level rather than repeated for every queried coordinate, improving both computational and memory efficiency.

A different but related direction is taken by Instant Neural Graphics Primitives (Instant-NGP) \cite{muller2022instant}, which introduces multiresolution structure at the level of the input encoding. Rather than modifying the network architecture itself or explicitly decomposing the signal into separate components, Instant-NGP equips a small neural network with a multilevel encoding of trainable feature vectors defined over grids of increasing resolution. For an input coordinate \(x\), each level produces an interpolated feature vector, and the outputs of all levels are concatenated and passed to the decoder MLP. Formally, this can be written as \(f(x)=g_\theta(\phi_1(x),\ldots,\phi_L(x))\), where \(\phi_\ell(x)\) denotes the interpolated feature vector at level \(\ell\). At coarse resolutions, grid vertices can be mapped one-to-one to stored entries, whereas at finer resolutions the feature vectors are stored in fixed-size hash tables, so multiple grid vertices may alias to the same entry. Instant-NGP does not explicitly resolve such collisions through probing or chaining; instead, it relies on gradient-based optimization to store appropriate sparse detail in the table and on the subsequent neural network to disambiguate collisions. It is important to highlight that the multiresolution design is crucial here, coarse levels are collision-free and capture low-resolution structure, while finer levels capture small features despite collisions, which are pseudo-randomly scattered across space and unlikely to occur simultaneously across all levels. When colliding samples are updated during training, their gradients accumulate, so samples with larger gradients tend to dominate, causing the representation to prioritize the sparse regions containing the most important fine-scale detail. In this way, Instant-NGP achieves a compact and highly efficient representation that captures both global structure and fine detail while enabling the use of a very small network. 

Taken together, these developments show that the power of INRs depends not only on the atoms they generate, but also on how those atoms are organized. Early models relied largely on global nonlinear expansions, whereas later designs imposed structure through scale, locality, and resolution. This reflects a broader shift from single global approximators to structured representations with complementary components.

\subsection{INRs Across Signal Modalities}
\label{sec:video}

We now broaden the discussion beyond static images, where the behavior of INRs becomes especially revealing because different modalities impose distinct structural demands. Audio emphasizes oscillatory and phase-sensitive structure, video introduces spatiotemporal redundancy, hyperspectral imaging extends the domain to continuous spectral signals, and 3D geometry introduces continuous spatial fields such as occupancy or signed distance functions. Together, these settings help clarify what a continuous representation must capture in order to model real-world signals faithfully.

\paragraph{\textcolor{blue}{Audio}}
Audio signals are inherently oscillatory and are dominated by fine-scale temporal variations. A waveform \(s(t)\) can often be expressed locally as a superposition of oscillatory components \( s(t) = \sum_k a_k(t)\cos(\omega_k t + \phi_k) \), where both amplitude and phase evolve over time. Unlike natural images, where energy often concentrated in low-frequency structures, audio demands precise modeling of high-frequency content and phase synchrony. This characteristic provides a rigorous stress test for coordinate-based networks, which inherently exhibit a spectral bias toward low-frequency approximations. Periodic activations provide a direct mechanism to represent such signals. In SIREN, each layer takes the form \( h^{(\ell+1)} = \sin\!\big(\omega_\ell (W^{(\ell)}h^{(\ell)} + b^{(\ell)})\big) \), introducing oscillatory basis functions directly into the representation. When modeling higher-frequency signals such as audio, larger values of $\omega_0$ in the first layer are often required in practice. The SIREN paper emphasizes that the first-layer frequency scaling controls the range of frequencies that the network can represent, although it does not prescribe a specific value as a function of signal bandwidth. In particular, SIREN for audio signals uses a substantially larger first-layer scaling for audio (e.g., $\omega_0 = 3000$), reflecting the fact that audio varies much more rapidly over the normalized input domain than typical image signals. As increasing $\omega_0$ enables the network to represent more rapidly varying components over the input coordinate, so the first layer effectively determines the accessible frequency range of the representation. Beyond representation, audio INRs also expose a limitation of per-signal INR fitting. Training a separate network for each waveform is computationally expensive, which motivates amortized approaches such as HyperSound \cite{szatkowski2022hypersound}, where a hypernetwork predicts INR parameters from input audio features.  This shifts the problem from optimizing a function per signal to learning a family of functions, effectively treating INRs as elements of a parameterized signal model. Subsequent work has explored compression-oriented formulations, where compact INR parameterizations achieve high-fidelity reconstruction, further reinforcing the interpretation of INRs as continuous codecs for oscillatory signals \cite{lanzendorfer2023siamese}.

\paragraph{\textcolor{blue}{Video}}

\begin{figure*}[t]
    \centering
    \includegraphics[width=\linewidth]{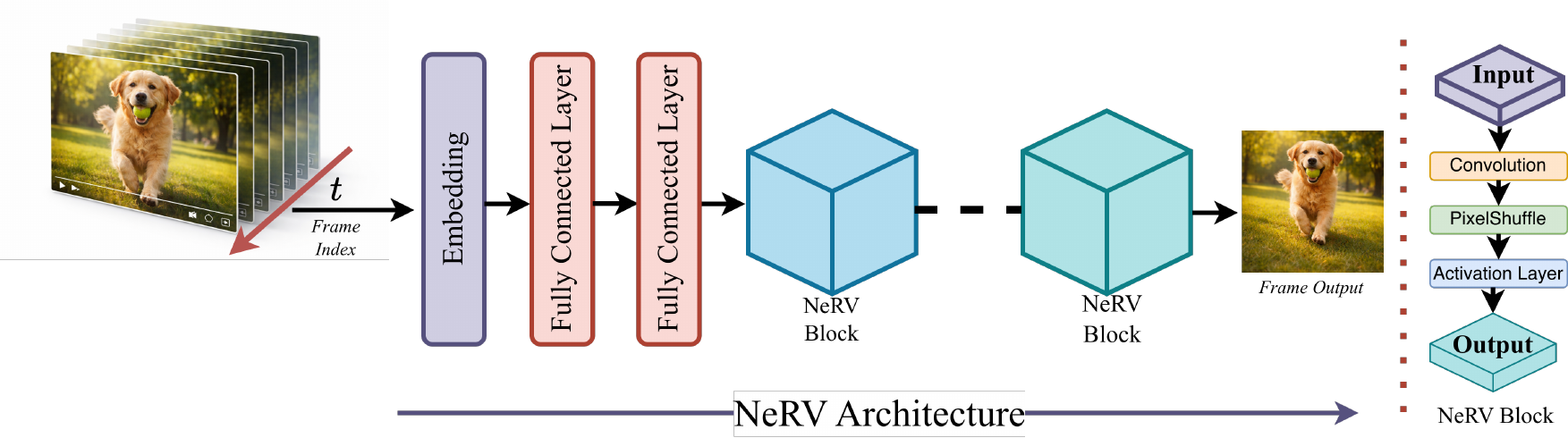}
\caption{Illustration of NeRV for video representation. A temporal index $t$ is mapped through embedding, fully connected layers, and NeRV blocks to synthesize the frame. Unlike standard INRs that evaluate every spatial coordinate, NeRV generates the full frame from a compact temporal input, enabling efficient representation.}\label{fig:nerv}
\end{figure*}
In contrast to audio, video signals introduce an additional layer of complexity through the coupling of space and time. A video may be viewed abstractly as a function \(f(x,y,t)\), where temporal redundancy plays a central role, adjacent frames are strongly correlated, much of the signal energy varies slowly over time, and only localized regions exhibit rapid motion or fine-scale detail. This immediately raises a representational question. While the pointwise formulation \(f(x,y,t)\) is a fully general formulation, it may not always the most natural parameterization for video \cite{chen2021nerv}. If every spatiotemporal sample is queried independently, the model must repeatedly reconstruct the same spatial organization across millions of pixel locations, even though most of that structure is shared within a frame and changes only gradually across time. NeRV \cite{chen2021nerv} addresses this by changing both the input parameterization and the synthesis mechanism. Instead of learning a pointwise mapping from coordinates to pixel intensities, it learns a framewise mapping \(f_\theta(t)=I_t\), where the input is the frame index or timestamp and the output is the entire image. This architectural shift is significant as the model no longer behaves like a coordinate-wise interpolator; it becomes a generator of frames. Equally important, once the output is a full frame, the representation is no longer restricted to the standard INR design of an MLP queried at individual coordinates, and the architectural diagram is shown in \Cref{fig:nerv}. NeRV introduces dedicated decoding blocks with convolutional structure, reshaping, and upsampling operations, so that spatial correlations are synthesized jointly rather than pixel by pixel. In effect, temporal variation is absorbed into a compact latent representation indexed by \(t\), while the spatial organization of each frame is reconstructed through convolutional feature generation. This architectural change is precisely what makes the NeRV family particularly well suited to video representation and compression-oriented settings. Subsequent works refined this idea by incorporating multiscale and adaptive mechanisms. HNeRV \cite{chen2023hnerv} introduced content-adaptive embeddings that redistribute representational capacity based on spatial complexity, while HiNeRV \cite{kwan2023hinerv} further improved compression efficiency through hierarchical encoding and quantization strategies. These developments highlight an important principle, video representations benefit from separating coarse structure from fine detail, both spatially and temporally. Overall, video highlights the need for representations that are not only expressive but also well structured. The challenge is not merely to capture high-frequency content, but to organize the representation so that redundancy across both space and time can be effectively exploited. 


\paragraph{\textcolor{blue}{Hyperspectral Imaging}}
Hyperspectral imaging extends INR modeling by introducing a spectral coordinate in addition to spatial position. A hyperspectral scene can be viewed as a function \(f(x,y,\lambda)\), where \(\lambda\) denotes wavelength. Unlike RGB images, which record only three channels, hyperspectral data samples the spectral radiance at many wavelengths, producing a high-dimensional signal with strong inter-band correlation. Let \(p=(x,y)\) denote a spatial location, and let \(R(p,\lambda)\) denote the spectral radiance at position \(p\) and wavelength \(\lambda\), for \(\lambda \in \Lambda\). A standard RGB sensor does not observe \(R(p,\lambda)\) directly; instead, each color channel \(c\in\{R,G,B\}\) records a weighted projection \(X_c(p)=\int_{\Lambda} R(p,\lambda)\Phi_c(\lambda)\,d\lambda\), where \(\Phi_c(\lambda)\) is the spectral response of the \(c\)-th channel. Zhang et al.~\cite{zhang2022implicit} use this observation to motivate INR-based hyperspectral super-resolution: the underlying spectrum is continuous, whereas conventional reconstruction methods estimate only a fixed discrete set of spectral bands.

Based on this idea, Zhang et al.~\cite{zhang2022implicit} introduced an INR formulation for hyperspectral super-resolution in which the network predicts the spectral signature at each spatial location. Their model takes the pixel coordinate together with the RGB value at that location and learns a mapping of the form \(f_\theta(X_p,p)\rightarrow \hat{Y}(p)\), where \(X_p\in\mathbb{R}^3\) is the observed RGB vector and \(\hat{Y}(p)\in\mathbb{R}^L\) is the reconstructed hyperspectral vector. Importantly, this is not a pure coordinate-only INR: the coordinate \(p\) is combined with measured image content, and the MLP parameters are predicted by a hypernetwork, making the representation content-adaptive. They further use periodic spatial encoding \( \gamma(p) \) so that the network can better recover high-frequency spatial details. A limitation of this formulation, however, is that the spectral dimension still appears only in the output as a discrete vector of \(L\) bands. Chen et al.~\cite{chen2023spectral} address this more directly in their spectral-wise implicit neural representation (SINR) for hyperspectral reconstruction under CASSI imaging \cite{chen2023spectral}. Instead of fixing the output to a predetermined band count, they formulate the reconstruction as a continuous spatial--spectral mapping, effectively learning a function over spectral coordinates as well. In their setting, the reconstruction network first extracts latent features from the coded measurement, and the implicit decoder then maps latent features together with spectral coordinates to reconstructed spectral values. This allows the model to reconstruct hyperspectral signals at arbitrary spectral resolutions rather than being tied to a fixed number of bands. The change in formulation is important. In the model of Zhang et al.~\cite{zhang2022implicit}, the INR primarily provides a continuous spatial representation whose output is spectrally high-dimensional. In SINR~\cite{chen2023spectral}, the INR is pushed further toward genuine spectral continuity: the decoder is asked to model variation with respect to wavelength itself. Chen et al.~\cite{chen2023spectral} further introduce spectral-wise attention to capture long-range inter-band dependencies and use Fourier coordinate encoding to mitigate spectral bias, since coordinate-based MLPs otherwise tend to favor smooth spectral variations and may miss fine spectral structure.

\paragraph{\textcolor{blue}{3D Geometry}}

\begin{figure}[t]
    \centering
    \includegraphics[width=\linewidth]{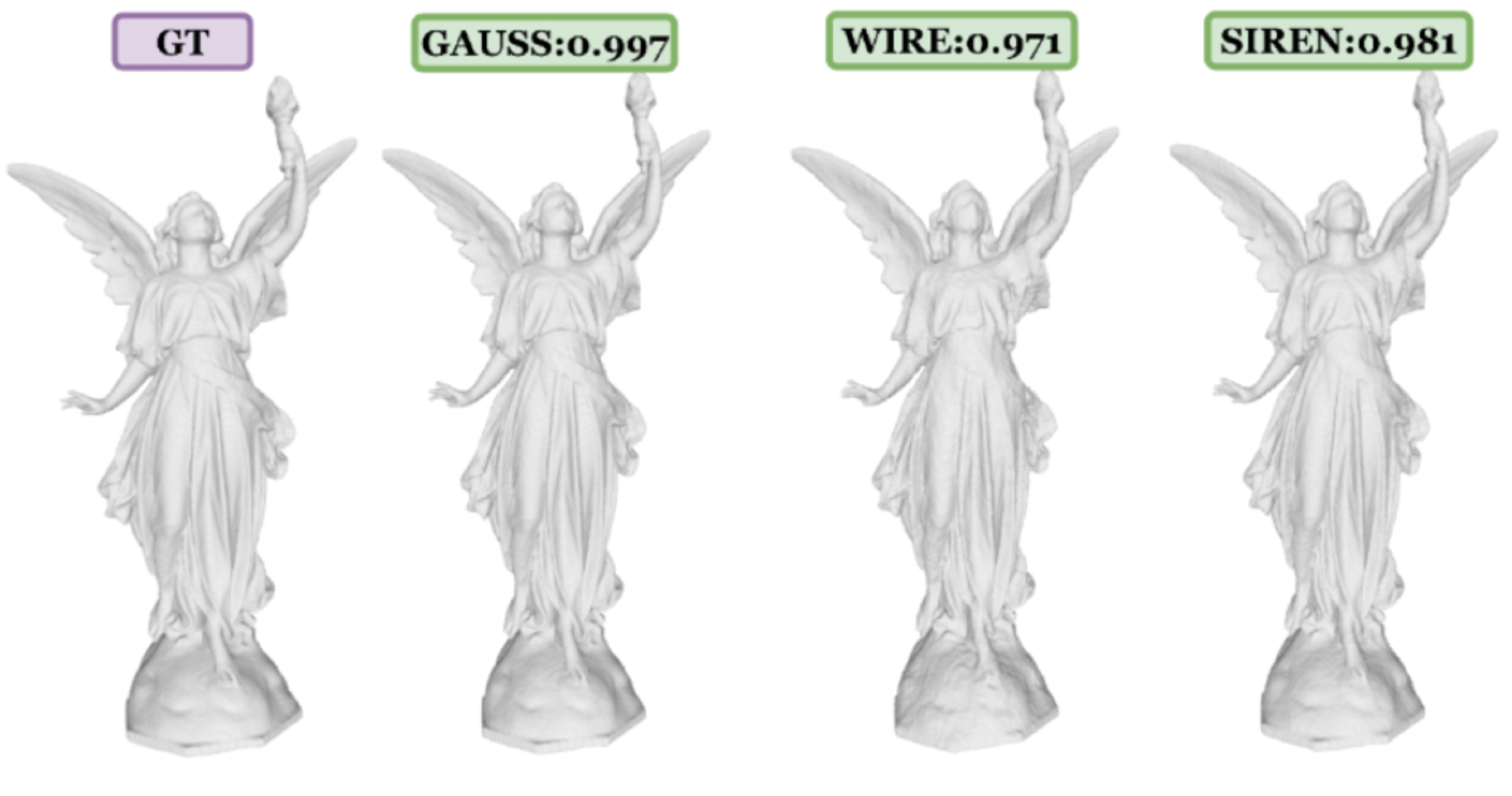}
    \caption{INR decoding of a complex 3D object. The reconstructions demonstrate how coordinate-based models capture fine geometric details and smooth surface structures in a continuous domain.}
    \label{fig:occupancy}
\end{figure}

A central question in INRs is how geometry should be encoded as a continuous signal. While all implicit models represent shapes as functions over spatial coordinates, they differ fundamentally in the type of signal assigned to each point. This choice directly determines the geometric information available to the model, as well as its ability to represent different classes of shapes \cite{mello2025neural}. An occupancy field models a function \(f_{\theta}(\mathbf{x}) \in [0,1]\) that indicates whether a point \(\mathbf{x} \in \mathbb{R}^{3}\) lies inside or outside the object, so that the surface is recovered as a decision boundary of the learned field (see \ref{fig:occupancy}). This formulation is intuitive and effective for continuous shape representation, but it provides only a binary notion of geometry and does not explicitly encode how far a point is from the surface. Signed distance functions (SDFs) enrich this representation by assigning each point a continuous value whose magnitude equals the closest distance to the surface and whose sign specifies whether the point lies inside or outside. As a result, the surface is defined by the zero level set \(f_{\theta}(\mathbf{x})=0\), while the field itself carries substantially richer geometric information. In particular, spatial derivatives of the SDF provide access to local geometric structure, such as surface normals, making SDFs especially attractive for reconstruction and rendering. However, this inside--outside convention also restricts SDFs mainly to closed, watertight surfaces. Unsigned distance functions (UDFs) relax this constraint by removing the sign and representing only the distance to the surface. This enables modeling of open or non-watertight geometries, but the loss of orientation information introduces ambiguity in surface extraction and weakens the geometric structure available to the model. In this sense, occupancy fields, SDFs, and UDFs reflect a progression in representational richness and flexibility: occupancy fields provide coarse inside--outside classification, SDFs offer a more informative geometric signal for closed surfaces, and UDFs trade some of that structure for the ability to model more general shapes.

Taken together, these examples show that there is no single INR architecture suited to all signal modalities. Their effectiveness depends on how well the parameterization and structural priors match the signal of interest. Audio often favors explicitly oscillatory models, video may benefit from framewise generation with convolutional decoding rather than pointwise spatiotemporal querying, and hyperspectral imaging requires representations that capture spatial-spectral coupling and spectral continuity. INR design is therefore inherently modality dependent.Our goal here has been to convey the central ideas behind these modality-specific formulations rather than to survey them exhaustively, and we encourage interested readers to explore the broader literature for further developments.

\subsection{Efficient evaluation and scalable INR architectures}
\label{sec:efficient}
As discussed, a persistent limitation of standard INRs is that each signal, image, shape, or scene is usually modeled by training a separate coordinate network from scratch. Even though this per-instance optimization yields a compact continuous representation, it is computationally expensive and shares little structure across related signals. It is also weak under sparse or incomplete observations, as optimization from a random initialization may fit the observed samples yet generalize poorly to unseen coordinates. 

\begin{wrapfigure}{r}{0.75\linewidth} 
    \centering
    \vspace{-10pt} 
    \includegraphics[width=\linewidth]{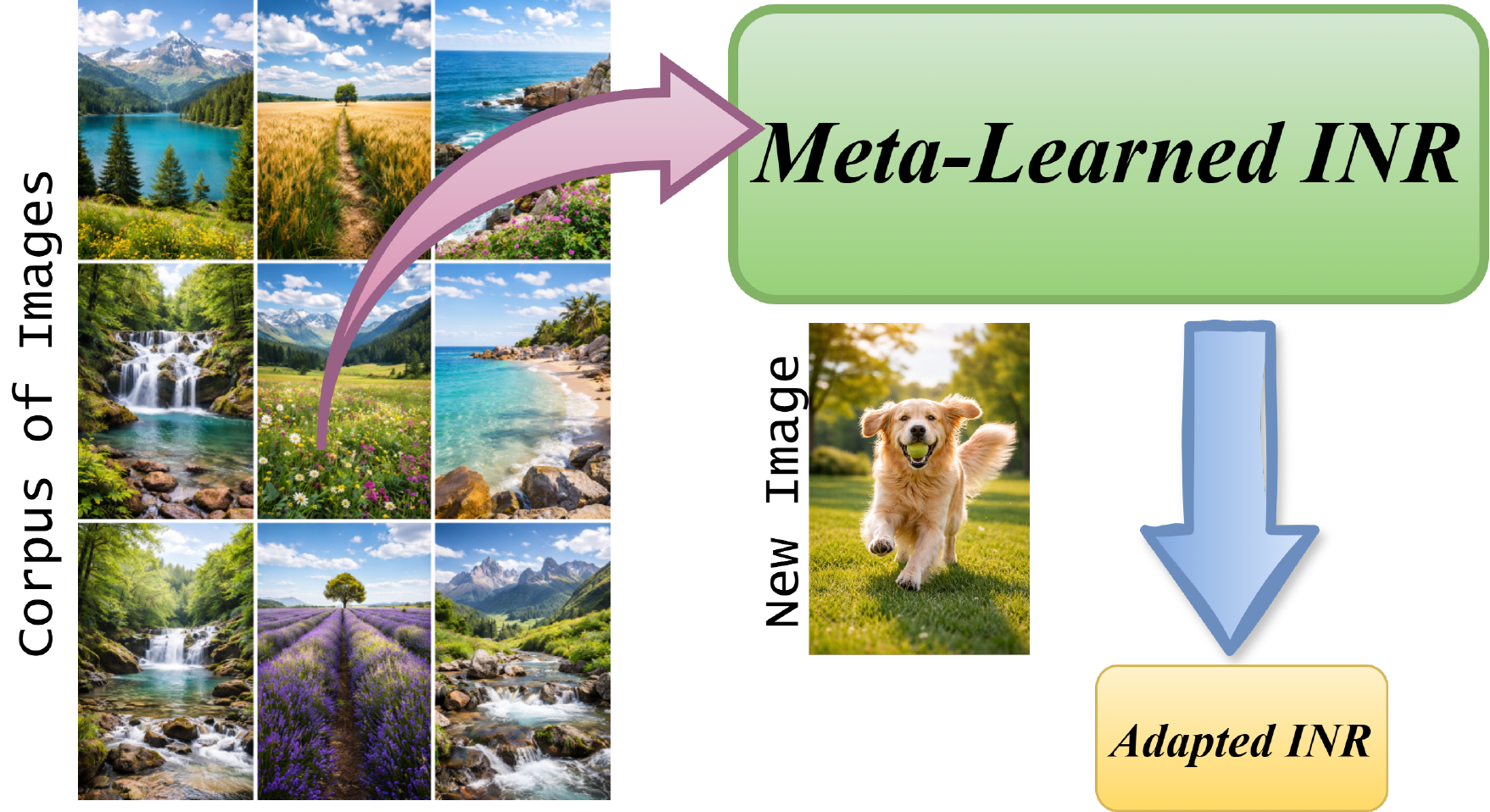}
\caption{Meta-learning for INRs. A meta-learned INR captures shared structure within a signal family, enabling rapid adaptation to new instances with minimal optimization.}
    \label{fig:meta_learned}
    \vspace{-10pt} 
\end{wrapfigure}

Meta-learning addresses this weakness by recasting INR fitting not as independent optimization for each signal, but as learning a prior over a family of functions. In MetaSDF \cite{sitzmann2020metasdf}, for example, each shape-specific signed distance function is treated as a task, and a shared meta-initialization is learned so that only a few gradient steps are needed to specialize it to a new shape. Concretely, let $\theta$ denote the shared meta-parameters learned across tasks, and let $\phi_i^j$ denote the task-specific INR parameters for task $i$ after the $j$-th inner-loop update. Further, let $X_i^{\mathrm{train}}$ denote the context set for task $i$, where each pair $(x,s)\in X_i^{\mathrm{train}}$ consists of a coordinate $x$ and its corresponding target signal value $s$. Let $\Phi(\cdot;\phi_i^j)$ denote the INR parameterized by $\phi_i^j$, let $\mathcal{L}$ denote the reconstruction loss, and let $\alpha$ denote the inner-loop learning rate. The adaptation step can then be written as $\phi_i^{j+1}=\phi_i^j-\alpha\nabla_{\phi_i^j}\sum_{(x,s)\in X_i^{\mathrm{train}}}\mathcal{L}(\Phi(x;\phi_i^j),s)$, with initialization $\phi_i^0=\theta$. This is a significant step as it replaces the latent-code bottleneck with a learned initialization in the full parameter space of the INR. MetaSDF shows that this yields performance competitive with auto-decoder methods while substantially reducing test-time inference cost \cite{sitzmann2020metasdf}. The underlying idea is inherently modality-agnostic and can be transferred naturally across signal domains; its adaptation to images is illustrated in \Cref{fig:meta_learned}. In the INR literature, this meta-learning paradigm has been applied to diverse tasks, including compression, audio representation, and quality assessment \cite{strumpler2022implicit,jayasundara2025sinr,jayasundarainriq,dupont2022coin++}, where learning a shared prior over function space improves both rapid adaptation to new signals and the efficiency of the resulting representations.

A natural next step is to move beyond fast adaptation and ask whether the parameters of an INR can be predicted directly, without requiring per-instance iterative optimization at inference time. This is the central idea behind hypernetworks. Instead of learning only an initialization that is later adapted, a hypernetwork learns a mapping from observations or conditioning variables to the parameters of a target INR~\cite{hyperinr_project2021}. Formally, if $\Phi(x;\theta)$ denotes the INR, a hypernetwork $H$ produces its parameters (or modulates them) as $\theta = H(z)$, where $z$ represents observations, latent variables, or task-specific conditioning. In this formulation, the INR need not be optimized separately for each signal, but can be synthesized in a single forward pass, effectively turning representation learning into a conditional function generation problem. This perspective can be interpreted as learning a mapping from signals to functions: rather than fitting $\Phi$ independently to each signal, the hypernetwork learns how signals are distributed across a family of functions and directly predicts the corresponding representation. In this sense, hypernetworks replace iterative reconstruction with a learned synthesis operator that generates continuous representations conditioned on input data. A representative example is HyperINR~\cite{wu2023hyperinr}, which uses a hypernetwork to generate a compact INR conditioned on scene or simulation parameters, rather than optimizing a separate representation from scratch for each instance. The core idea is to combine shared structure with condition-specific modulation, yielding efficient representations that can also generalize to unseen settings. Taken together, these developments mark a clear progression: meta-learning learns how to initialize and adapt INRs efficiently, whereas hypernetworks learn how to synthesize the representation itself.

\section{Downstream tasks and task-oriented inference with INRs}

\subsubsection{\textcolor{blue}{Medical imaging applications of INRs}}

Medical imaging is a particularly natural setting for INRs because many of its core tasks are, at heart, continuous-domain inverse problems. Anatomical structures, physiological processes, and motion are inherently continuous, while the measurements available in practice are often sparse, noisy, anisotropic, motion-corrupted, or only indirectly related to the underlying object through an imaging system \cite{molaei2023implicit}. This gap makes the INR perspective especially compelling. Rather than regarding the sampled voxel grid as the true object of interest, INRs model the underlying image, deformation field, or anatomy itself as a continuous coordinate-based function. From a signal-processing perspective, this brings the formulation closer to the real problem: recovering a continuous signal through a forward model, while regularizing the solution with a structured continuous prior.

This viewpoint is especially clear in reconstruction. For example, \emph{NeSVoR} models the target volume as a continuous function together with a continuous acquisition model that accounts for slice motion, point-spread effects, and bias fields, thereby casting slice-to-volume MRI reconstruction as estimation of a latent continuous image under measurement physics \cite{xu2023nesvor}. A similar advantage appears in registration, where the object to be estimated is itself a continuous deformation field rather than a discrete displacement grid. In this setting, implicit parameterizations make smoothness and derivative-based regularization more natural to impose. Segmentation reveals a different but equally important role for INRs. Earlier work such as \emph{IOSNet} showed that implicit segmentation can provide memory-efficient, resolution-agnostic prediction \cite{khan2022implicit}. More recently, \emph{MetaSeg} demonstrated that INRs can also be made competitive for predictive segmentation by meta-learning an initialization that adapts to unseen images while simultaneously decoding image intensity and class labels. This is important because it addresses a core weakness of standard INRs: although they are effective as per-instance continuous models, they are not naturally designed to learn semantic structure across a population of images \cite{vyas2025fit}. Taken together, these examples suggest that INRs are valuable in medical imaging not simply because they are continuous, but because continuity aligns naturally with the way anatomy, acquisition, deformation, and supervision are modeled. For a broader overview of INRs in medical imaging, we refer interested readers to the survey by Molaei \emph{et al.} \cite{molaei2023implicit}.

\subsubsection{\textcolor{blue}{Radar imaging}}

Radar imaging is a particularly compelling application for INRs as the inverse problem is inherently sparse, indirect, and physics-governed. In 3D Synthetic Aperture Radar (SAR), the sensor does not observe a dense volumetric image directly; rather, it measures incomplete slices of the scene’s 3D spatial Fourier domain \cite{sugavanam2026neural}. In practice, apertures are often sparse or non-uniform, especially in elevation, which leads to ill-posed inversion, height ambiguities, aliasing, and scattered point-cloud reconstructions when classical Fourier inversion or sparsity-based voxel methods are used. Sugavanam and Ertin \cite{sugavanam2026neural} argue that, in this regime, a continuous implicit model is preferable to a fixed voxel grid, as the dominant SAR returns often arise from object surfaces rather than filled volumes. They therefore represent the scene by a neural signed distance function, $\mathrm{SDF}(\mathbf{p})$, whose zero level set captures the scattering surface, thereby regularizing sparse and noisy reconstructions with a geometric prior rather than only an $\ell_1$ prior on discrete voxels.  Instead of representing radar reflectivity on a fixed voxel grid, INR-based methods model the scene as a continuous function over spatial coordinates, or, in surface-oriented formulations, as an implicit surface whose zero level set describes the object boundary. This is attractive for radar because it replaces staircase-like voxel artifacts with a continuous spatial prior that can be queried at arbitrary resolution.

A second motivation is that radar measurements are strongly view- and geometry-dependent, so the representation should be coupled to the acquisition model. In SAR target recognition, Cheng \emph{et al.} \cite{cheng2025implicit} note that data augmentation or black-box generative synthesis does not explicitly encode SAR imaging geometry and therefore lacks interpretability. Their SAR-INR instead models the scene backscatter as a continuous function of spatial location and radar beam incidence angle,
$
(p,\sigma)=F(X,Y,Z,\theta,\phi),
$
where $(X,Y,Z)$ denotes the queried 3D point and $(\theta,\phi)$ the radar line of sight. This representation is tied to a SAR projection model that maps world coordinates to image coordinates, allowing the network to be trained through a differentiable rendering pipeline. As a result, unseen viewpoints can be synthesized by querying the same learned field under new imaging angles, which is especially valuable when labeled SAR views are limited. 

Recent work on Frequency-Modulated Continuous-Wave (FMCW)  radar further shows that the benefit of INRs is not only continuity, but also better alignment with radar signal formation \cite{takawale2025spinrv2}. SpINRv2 \cite{takawale2025spinrv2} observes that range is linearly encoded in beat frequency,
so supervision in the frequency domain is more natural than supervising raw time-domain beat signals. Time-domain learning can be poorly conditioned 
because small geometric perturbations induce large phase shifts, while coarse range-quantized models ignore phase interactions, spectral leakage, and sub-bin effects. SpINRv2 therefore combines a continuous INR scatterer field with a differentiable frequency-domain forward model, and augments the spectral loss with smoothness and sparsity priors,
$
\mathcal{L}_{\text{total}}=\mathcal{L}_{\text{spectral}}+\beta \mathcal{L}_{\text{smooth}}+\gamma \mathcal{L}_{\text{sparsity}}.
$
This formulation preserves amplitude and phase information, and yields more accurate volumetric reconstructions under high-frequency sensing conditions. 
Taken together, these works suggest that INRs are well matched to radar as they provide a continuous scene model that can absorb geometric priors, surface priors, and signal-domain physics within a single differentiable framework. Rather than treating radar as just another image modality, they model what radar actually measures: view-dependent scattering shaped by aperture geometry, wave propagation, and frequency-domain structure. 

\subsubsection{\textcolor{blue}{Compression with INRs: coding continuous decoders}}

Compression is one of the clearest settings in which the distinct nature of INRs becomes apparent. Classical source coding transforms a signal into a domain where redundancy is easier to remove, and then applies quantization and entropy coding under a rate-distortion objective. JPEG, for example, represents image blocks with DCT coefficients before quantization and entropy coding \cite{smith1997scientist}. INR-based compression retains this overall logic, but shifts compression to the parameter space of a continuous coordinate-to-value function \(f_{\boldsymbol{\theta}}\). Instead of coding sampled values or transform coefficients on a fixed grid, it transmits a coded description of the learned parameters \(\boldsymbol{\theta}\) \cite{dupont2021coin, strumpler2022implicit}. The resulting bitstream therefore specifies a continuous decoder that can reconstruct the signal at arbitrary coordinates through \(f_{\boldsymbol{\theta}}(\mathbf{x})\).

This principle was made explicit by \emph{COIN} \cite{dupont2021coin}, which compresses an image by overfitting a coordinate-based network and transmitting its quantized weights. The importance of this idea is conceptual as much as practical. Once the decoder is a continuous function, reconstruction is no longer tied to a fixed lattice; the same code naturally supports arbitrary querying, nonuniform sampling, and resolution-flexible decoding. At the same time, COIN also exposes a central difficulty of INR-based compression: naively training a separate INR for each image and transmitting its full parameter vector is expensive, while direct weight quantization does not exploit structure shared across signals.  This is precisely where \emph{COIN++} \cite{dupont2022coin++} refines the picture. Instead of learning and transmitting a separate full INR for each signal, it learns a shared decoder \(f_{\boldsymbol{\theta}}(\mathbf{x};\boldsymbol{\phi})\), where \(\boldsymbol{\theta}\) denotes parameters common to the whole dataset and \(\boldsymbol{\phi}\) is a low-dimensional code specific to a single signal. Compression is therefore achieved by transmitting only \(\boldsymbol{\phi}\) for each instance, while the shared decoder remains fixed. The coded object thus changes in an important way: rather than sending an entire continuous decoder for every image, one sends only a compact signal-specific code that adapts a common decoder to the source instance. From a coding standpoint, common structure is absorbed into the shared base network, while the transmitted bits carry only the source-specific residual variation. 

Another line of work is motivated by a classical principle in signal compression, many natural signals admit compact representations because they are sparse, or approximately sparse, in an appropriate transform domain. Rather than viewing the INR only as a continuous decoder for the signal, SINR\cite{jayasundara2025sinr} goes one step further and asks whether the learned INR parameters themselves may also possess such transform-domain sparsity. Concretely, if \(\mathbf{w}\in\mathbb{R}^P\) denotes the trained parameter vector of the INR for a hidden layer, SINR seeks a sparse approximation of the form \( \mathbf{w}\approx A\boldsymbol{\alpha} \), where \(A\) is a dictionary and \(\boldsymbol{\alpha}\) is a sparse coefficient vector satisfying a constraint such as \( \|\boldsymbol{\alpha}\|_0\le s \). Compression is then achieved by quantizing and coding only the nonzero coefficients in the sparse code, together with their indices, rather than the full parameter vector. The key idea is therefore that the INR is not only a continuous representation of the signal; it can itself be treated as an object that may admit a compact transform-domain description. In this sense, SINR introduces an additional layer of compression, moving from sparsity of the signal to sparsity of the continuous decoder used to represent it. These developments suggest that INR-based compression treats the coded object as a continuous reconstruction model rather than merely a sampled signal. This leads to a more unified view of coding across signal modalities, while still allowing the model design to adapt to the underlying structure of the data.

\subsubsection{\textcolor{blue}{Scene representation}}

One of the most influential downstream uses of INRs is the neural radiance field (NeRF), introduced by Mildenhall et al.~\cite{mildenhall2021nerf}. The task is novel-view synthesis: given a set of posed images of a scene, the goal is to learn a representation that can render the same scene from viewpoints not observed during training. This is a more demanding problem than fitting a single image or waveform, because the representation must account for geometry, appearance, and visibility simultaneously. 

\begin{figure}[t]
    \centering
    \includegraphics[width=\linewidth]{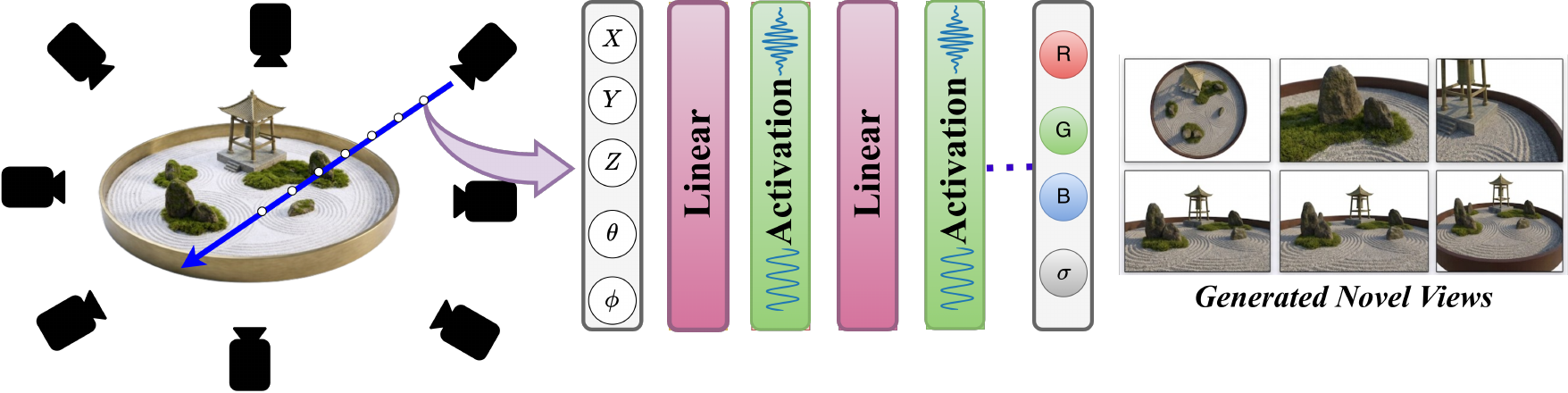}
    \caption{NeRF represents a scene as a continuous function of spatial location and viewing direction, enabling reconstruction of geometry and appearance and high-quality novel view synthesis.}
    \label{fig:novel_views}
\end{figure}

Earlier approaches often relied on explicit intermediate representations such as meshes, point clouds, or voxel grids. These can be effective, but they also introduce familiar difficulties: geometry estimation can be brittle, voxel methods are memory intensive and resolution limited, and image-based schemes may struggle to maintain consistency across viewpoints \cite{mildenhall2021nerf}. NeRF replaces these explicit structures with a continuous scene model learned directly from images. More precisely, NeRF represents a scene by a neural function \(F_{\boldsymbol{\theta}}(\mathbf{x},\mathbf{d})=(\sigma,\mathbf{c})\), where \(\mathbf{x}=(x,y,z)\in\mathbb{R}^3\) is a 3D spatial coordinate and \(\mathbf{d}=(\theta,\phi)\in\mathbb{S}^2\) is a viewing direction. Here, \(\sigma\in\mathbb{R}_{+}\) denotes the volume density and \(\mathbf{c}\in\mathbb{R}^3\) denotes the emitted RGB radiance. Intuitively, \(\sigma\) captures where matter is likely to exist in the scene, whereas \(\mathbf{c}\) specifies how that point appears from the direction \(\mathbf{d}\). Therefore, the INR is no longer just approximating signal values over coordinates; it is encoding a volumetric scene whose outputs are tied directly to image formation. Training is performed through differentiable rendering \cite{kato2020differentiable}. For each pixel in a training image, one casts a camera ray through the scene, queries the field \(F_{\boldsymbol{\theta}}\) at sampled points along that ray, composites the resulting colors and densities into a rendered pixel color \(\hat{C}(\mathbf{r})\), and compares it with the observed color \(C^{\mathrm{gt}}(\mathbf{r})\) (See \Cref{fig:novel_views}). The model is optimized with a reconstruction objective such as \(\mathcal{L}_{\mathrm{rec}}=\sum_{\mathbf{r}\in\mathcal{R}}\|\hat{C}(\mathbf{r})-C^{\mathrm{gt}}(\mathbf{r})\|_2^2\), where \(\mathcal{R}\) denotes a batch of training rays. The key point is that supervision is imposed at the image level, but the learned object is a continuous 3D field. Because the rendering process is differentiable, image errors can be propagated back to update the scene representation itself. A central practical challenge is that scenes contain fine geometric and photometric detail, whereas coordinate-based MLPs tend to prefer smooth, low-frequency solutions. NeRF addresses this by using positional encoding on both spatial coordinates and viewing directions, allowing the network to better recover high-frequency scene content. It also uses hierarchical sampling so that computation is concentrated in informative regions of space rather than wasted in empty volume. These two ingredients are not incidental; they are what make the continuous representation expressive enough to model detailed scenes from sparse image supervision. What makes NeRF especially significant in the INR literature is that it elevates the representation from a signal model to a scene model. Once the scene is encoded as \(F_{\boldsymbol{\theta}}(\mathbf{x},\mathbf{d})=(\sigma,\mathbf{c})\), the same representation can support not only novel-view synthesis, but also scene editing, inpainting, object removal, and other geometry-aware manipulations while preserving multi-view consistency \cite{mirzaei2023spin}. In this sense, NeRF demonstrates one of the clearest ways in which an INR can move beyond function fitting and become the computational object on which downstream vision and graphics tasks are performed. 

\subsubsection{\textcolor{blue}{INR-based classification: from weight vectors to function-space features}}

A useful aspect of INRs is that they decouple the underlying signal model from the specific grid on which the signal is observed. This gives classification a richer interpretation. Rather than relying only on sampled pixels or measurements, recognition can instead be performed on the continuous representation fitted to the data. In practice, discrimination may be based on the INR weights, the behavior of the learned function itself, or features extracted along its internal representation path. The central challenge, then, is to identify which level of the implicit model most effectively captures class-relevant structure. A natural starting point is to treat the fitted INR weights themselves as features. Malherbe~\cite{malherbe2024implicit} introduced this idea explicitly, showing that the parameter vector of a trained INR can be used as a representation for structured data such as sound, images, videos, or accelerometer signals, and then classified with standard models such as XGBoost. In this formulation, one first fits a continuous function \(f_{\boldsymbol{\theta}}\) to a signal and then predicts its class label from the corresponding parameter vector \(\boldsymbol{\theta}\). The attraction of this approach is its generality: once different modalities are mapped into INR parameter space, a common downstream classifier can in principle operate across all of them. At the same time, this weight-space view also exposes a limitation. The parameters of an INR are not a canonical description of the underlying function; different weight vectors may represent essentially the same signal. This means that geometric proximity in parameter space does not always align well with semantic similarity. In practice, direct weight classification is therefore attractive as a simple baseline, but it does not fully exploit the structure of the learned function.

\begin{figure}[t]
    \centering
    \includegraphics[width=\linewidth]{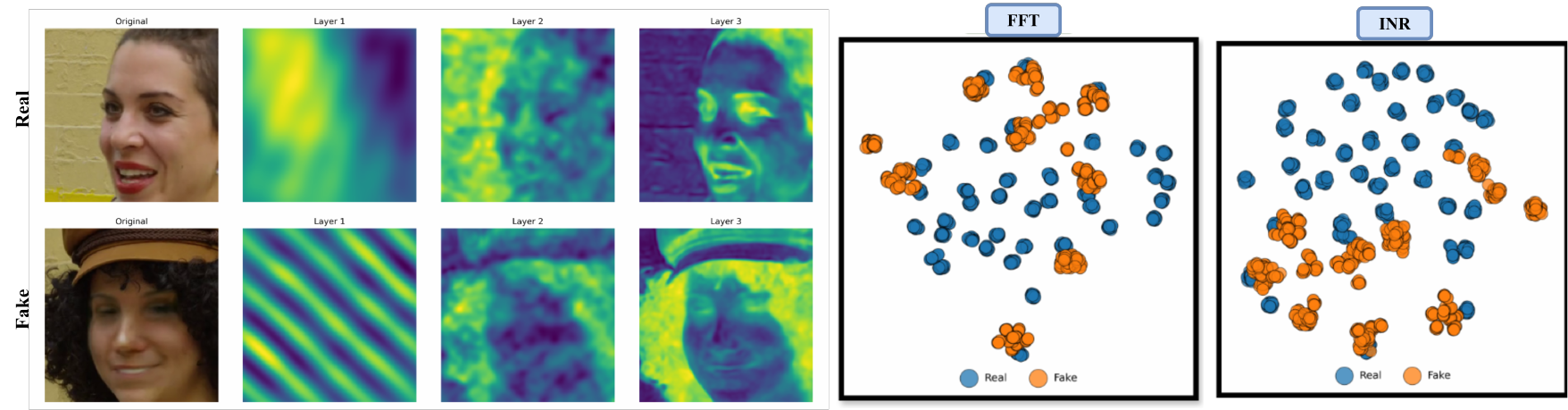}
\caption{INR-based feature representations for classification. Left: layer-wise INR heatmaps for real and fake samples, highlighting differences in spatial–frequency patterns. Right: t-SNE visualizations showing improved class separability when combining INR heatmaps with CLIP features compared to FFT-based representations.}\label{fig:deepfake}
\end{figure}

A more function-aware direction was introduced by Chen et al.~\cite{chen2022resolution}, who proposed classifying remote sensing scenes \emph{in function space} rather than from rasterized image inputs. Their key motivation is that CNN-based classifiers are often sensitive to image resolution, whereas an INR decouples resolution from the represented scene by modeling the image as a continuous function. In their approach, the image is first converted into an INR through optimization, and classification is then performed on the resulting function-space representation using a simple MLP classifier. This is an important conceptual shift: the classifier does not operate on pixels directly, but on a continuous representation whose behavior is less tied to a particular sampling grid.  Another viewpoint is to use the INR not merely as a fitted function, but as a generator of multiscale discriminative features. This idea is particularly evident in recent authenticity-related tasks. For instance, \emph{INFER} \cite{jayasundarainfer} leverages a fitted INR to extract layer-wise activation features, which are aggregated into heatmaps and fused with CLIP embeddings \cite{radford2021learning} for deepfake detection, enabling sensitivity to subtle structural artifacts (see \Cref{fig:deepfake}). A different route is taken by Gielisse and van Gemert, \cite{gielisse2025end} who move back to weight-space classification but make the INR fitting process itself task-aware: instead of first fitting a SIREN and then classifying its weights, they meta-learn the initialization and update dynamics so that the fitted parameters become more directly interpretable for a downstream Transformer classifier. In this way, classification feedback shapes the structure of the INR during fitting, improving discriminability without explicitly enforcing symmetry-equivariant architectures.

\subsubsection{\textcolor{blue}{Image quality assessment (IQA)}}

IQA seeks to quantify perceptual degradation rather than only pointwise numerical error. Classical measures such as MSE and PSNR compare images in the pixel domain, SSIM incorporates local structural information, and more recent learned metrics rely on deep features extracted by pretrained networks \cite{zhai2020perceptual}. INR-based IQA poses a different question, which is, can quality be inferred from how a signal is represented by an implicit model, rather than only from sampled intensities or external embeddings? As INR represents an image as a continuous function \(f_{\theta}\) that maps coordinates to intensities or colors, with the parameters \(\theta\) obtained by fitting the network to the signal. This immediately suggests a new viewpoint for IQA. Rather than comparing two images only as arrays, one may compare the representations induced by those images. A naïve version of this idea would be to compare the fitted parameters of two INRs directly. However, the INRIQ \cite{jayasundarainriq} work shows that this may not sufficiently stable: raw weight differences, and even simple activation-based comparisons, are strongly affected by optimization details and may not reliably track degradation severity. This motivates a more informative quantity: not the weights themselves, but how a reference-trained INR would need to change in order to explain a degraded signal. This is the key idea behind \emph{INRIQ}. Let \(f_{\theta_{\mathrm{ref}}}\) denote an INR trained on a reference image \(x_{\mathrm{ref}}\). Keeping this network fixed, one evaluates the loss of the reference INR on another signal \(x\) (a degraded version of original signal) and computes the gradient of that loss with respect to the reference parameters. These gradients describe how the reference representation would need to move, infinitesimally, to better fit the new signal. Aggregating the squared gradients over coordinates yields a Fisher-type sensitivity profile for the image. Quality is then assessed by comparing the sensitivity profiles induced by the reference and degraded images, using a symmetrized KL divergence.  This gives a degradation-sensitive notion of difference tied to the geometry of the learned model rather than only to pointwise intensity error.  A complementary direction appears in Beyond Pixels \cite{ozer2025beyond}, where the goal is not necessarily to compute a continuous similarity score, but to classify image quality into categories such as good, mid, and poor. In this setting, each image is first fitted by an INR, and the learned parameter vector itself is treated as a compact descriptor of image quality. The prediction problem is therefore shifted from pixel-space classification to weight-space classification, that is, from judging the image directly to judging the representation learned for that image. The reported results show that classifiers operating on INR parameters outperform a pixel-based MLP baseline, suggesting that degradation-relevant structure is naturally organized in the learned representation. An interesting observation is that even early-stage INR parameters already contain predictive information, although the best performance is obtained once the fit becomes sufficiently accurate. 


\section{Open Problems and Future Directions}
\label{sec:open_problems}

Despite recent progress, several important questions remain open regarding the role of INRs as signal-processing tools. Their continuous and differentiable formulation creates new opportunities for interpolation, inverse reconstruction, and physics-informed modeling, but it also raises challenges related to analysis, scalability, and interpretability.

One important direction concerns \emph{theoretical understanding}. Recent studies have provided useful insight into spectral bias and activation behavior \cite{saratchandran2024sampling,ramasinghe2022beyond}, but many broader questions remain unresolved. For example, it is still not fully understood under what conditions an INR can be expected to represent or recover a signal reliably, and how this depends on factors such as the signal class, sampling density, measurement operator, and architectural design. As a result, many INR formulations are still guided heavily by empirical practice. A more principled understanding, even in restricted settings, could help clarify when particular designs are appropriate and provide stronger insight into approximation, stability, and recovery behavior.

Another promising direction is a better understanding of the \emph{weight space} of INRs. Although a signal is ultimately encoded by the parameter vector \(\boldsymbol{\theta}\), it remains unclear how these parameters are organized across different signals. In particular, it is not yet well understood whether perceptually or structurally similar signals map to nearby regions of parameter space, or whether quite different parameter configurations can represent closely related functions. This question is further complicated by symmetries and redundancies in neural parameterizations. A better understanding of the geometry and invariances of INR weight space could eventually lead to more meaningful similarity measures between signals, and may also benefit applications such as compression, interpolation across representations, image generation, and downstream recognition.

Another major challenge lies in scalability and generalization. Many INR methods are still optimized per instance, limiting their applicability to large-scale datasets, long videos, or dynamic 3D environments. Future work should focus on amortized or meta-learned INRs that generalize across scenes and tasks, as well as architectures that can efficiently model spatiotemporal signals over long horizons. Additionally, integrating INRs with generative models (e.g., diffusion or transformer-based frameworks) offers a promising path toward controllable, high-fidelity synthesis. Finally, there is a growing need for physics-aware and uncertainty-aware INRs, particularly in scientific and medical applications, where incorporating domain constraints and providing reliable confidence estimates are critical for real-world deployment.

Beyond reconstruction and generation, INRs open up several under-explored application directions in perception and decision-making. These include tasks such as object detection, spatial localization, segmentation, and tracking, where continuous coordinate-based representations could offer higher precision and scale invariance. INRs also have potential in robotics and embodied AI, for example as world models or policy representations, as well as in multimodal settings for sensor fusion and cross-view understanding. Extending INRs to these tasks represents an important shift from modeling signals to enabling full perception, reasoning, and control pipelines.

\section{Conclusion}


Implicit neural representations (INRs) shift signal modeling from discrete, grid-bound samples toward continuous functional representations. By parameterizing signals with neural networks, they provide a unified framework for reconstruction, compression, and inverse problems. Throughout this article, we have argued that their behavior is closely connected to classical ideas from harmonic analysis, approximation theory, and sampling theory. From periodic activations to multiscale structured designs, the evolution of INR architectures can be viewed as a search for representations that balance spectral expressivity with spatial and temporal efficiency. In this sense, INRs are not simply neural architectures, but learned signal models whose approximation spaces adapt to the structure of the data. Looking ahead, their continued development will likely depend on closer integration between signal-processing insight and learning-based design. Viewed in this way, INRs are less a departure from signal processing than a learned continuous extension of it.

\bibliographystyle{ieeetr}  
\bibliography{main}

\vfill
\end{document}